\documentclass[10pt,journal]{IEEEtran}
\ifCLASSOPTIONcompsoc
  \usepackage[nocompress]{cite}
\else
  \usepackage{cite}
\fi
\ifCLASSINFOpdf
\else
\fi
\usepackage{cite}
\usepackage{amsmath,amssymb,amsfonts}
\usepackage{algorithmic}
\usepackage{textcomp}
\usepackage{multicol}
\usepackage{lipsum}
\usepackage{longtable}
\usepackage{vcell}
\usepackage{wrapfig}
\usepackage{algorithm}
\usepackage{epsfig}
\usepackage{graphicx}
\usepackage{booktabs}
\usepackage{multirow}
\usepackage{hhline}
\usepackage{textcomp}
\usepackage{url}
\usepackage[leftcaption]{sidecap}
\usepackage{caption}
\usepackage{xcolor, soul}
\usepackage{colortbl}
\usepackage{caption}
\usepackage{subfig}
\usepackage{amsthm}
\usepackage{etoolbox}

\usepackage{array}

\begin{document}
%
\title{Large Language Models for UAVs: Current State and Pathways to the Future}
\author{Shumaila Javaid \IEEEmembership{Member, IEEE}, Nasir Saeed \IEEEmembership{Senior Member, IEEE}, Bin He \IEEEmembership{Senior Member, IEEE}
\thanks{S. Javaid and B. He are with the Department of Control Science and Engineering, College of Electronics and 	Information Engineering, Tongji University, Shanghai 201804, China, and also with Frontiers 	Science Center for Intelligent Autonomous Systems, Shanghai 201210. Email: \{shumaila, hebin\}@tongji.edu.cn.\newline
N. Saeed is with the Department of Electrical and Communication Engineering, United Arab Emirates University (UAEU), Al Ain, 15551, UAE. Email: mr.nasir.saeed@ieee.org.}}


\maketitle

\begin{abstract}
Unmanned Aerial Vehicles (UAVs) have emerged as a transformative technology across diverse sectors, offering adaptable solutions to complex challenges in both military and civilian domains. Their expanding capabilities present a platform for further advancement by integrating cutting-edge computational tools like Artificial Intelligence (AI) and Machine Learning (ML) algorithms. These advancements have significantly impacted various facets of human life, fostering an era of unparalleled efficiency and convenience. Large Language Models (LLMs), a key component of AI, exhibit remarkable learning and adaptation capabilities within deployed environments, demonstrating an evolving form of intelligence with the potential to approach human-level proficiency. This work explores the significant potential of integrating UAVs and LLMs to propel the development of autonomous systems. We comprehensively review LLM architectures, evaluating their suitability for UAV integration. Additionally, we summarize the state-of-the-art LLM-based UAV architectures and identify novel opportunities for LLM embedding within UAV frameworks. Notably, we focus on leveraging LLMs to refine data analysis and decision-making processes, specifically for enhanced spectral sensing and sharing in UAV applications. Furthermore, we investigate how LLM integration expands the scope of existing UAV applications, enabling autonomous data processing, improved decision-making, and faster response times in emergency scenarios like disaster response and network restoration. Finally, we highlight crucial areas for future research that are critical for facilitating the effective integration of LLMs and UAVs.
\end{abstract}


\section{Introduction}
\label{sec:introduction}
Unmanned Aerial Vehicles (UAVs) have been a focus of attention for over fifty years due to their remarkable autonomy, mobility, and adaptability, enhancing a wide array of applications, including surveillance \cite{li2021networked,thakur2021artificial}, monitoring \cite{ren2019review,popescu2019survey}, search and rescue \cite{ Khalil2023677}, healthcare \cite{ullah2019uav}, maritime communications \cite{akhtar2023uavs}, and wireless network provisioning \cite{mozaffari2019tutorial}. These foundational achievements prompted the integration of Artificial Intelligence (AI) with UAVs. Particularly in the 2010s, advances in both UAV technology and AI reached a critical juncture, leading to substantial benefits across various applications. For example, AI-enabled UAVs employ facial recognition and real-time video analysis techniques to enhance security and monitoring of remote areas \cite{boroujeni2024comprehensive,koubaa2023aero,cheng2023ai}. In agriculture, UAVs with AI models analyze crop health for precise farming, improving resource efficiency and yields \cite{qazi2022iot,vincent2019sensors}. Meanwhile, AI-driven UAVs optimize logistics route planning and inventory management, streamlining warehouse operations and improving delivery efficiency \cite{iyer2021ai,al2020comprehensive, Khalil2024}.

Among these advancements, Large Language Models (LLMs) have recently gained significant attention as they enable systems to learn from application behavior and optimize existing systems \cite{mohammadi2018enabling,rasley2020deepspeed}. Various LLMs employing transformer architectures, such as the Generative Pre-trained Transformer (GPT) series \cite{ye2023comprehensive}, Bidirectional Encoder Representations from Transformers (BERT) \cite{devlin2018bert}, and Text-to-Text Transfer Transformer (T5) \cite{mastropaolo2021studying}, exhibit fundamental capabilities. Due to extensive training on large datasets, they excel at understanding, generating, and translating human-like text, making them valuable for robotics, healthcare, finance, education, customer service, and content creation applications.
Furthermore, the proficiency of these models in real-time data processing, natural language understanding and generation, content recommendation, sentiment analysis, automated response, language translation, and content summarization creates opportunities in the UAV domain. For instance, they enable UAVs to respond swiftly to dynamic environmental changes and communication demands \cite{liu2024generative,agapiou2023interacting}. Their adaptive learning capabilities facilitate continuous improvement in operational strategies based on incoming data, enhancing decision-making processes \cite{kurunathan2023machine}. Additionally, their ability to support multiple languages broadens their applicability in global operations, particularly valuable for UAV communications in diverse applications such as smart cities, healthcare, rescue operations, emergency response, media, and entertainment \cite{rong2024leveraging, ullah2024role,liu2023aerialvln}.

Recent literature \cite{piggott2023net,de2023socratic,sun2024generative} has explored incorporating LLMs into UAV communication systems to enhance interaction with human operators and among UAVs. Traditionally, UAVs operate on pre-programmed commands with limited dynamic interaction capabilities. However, integrating LLMs enables support for natural and intuitive communication methods. For example, LLMs can interpret and respond to commands in natural language, simplifying UAV control and allowing the handling of complex, real-time mission adjustments. This transforms UAVs into more adaptable and practical tools across various applications \cite{yun2021attention}.
LLMs can enhance UAVs' autonomous decision-making based on communication context or environmental data \cite{zhao2024expel,eigner2024determinants}. For example, without human input, LLMs can analyze messages and environmental data in search and rescue operations to determine priorities and actions. In multi-UAV operations, LLMs can facilitate better communication and coordination, managing and optimizing information flow between UAVs and improving overall efficiency and effectiveness. LLMs can also enhance data processing and reporting capabilities by generating summaries, insights, and actionable recommendations from vast amounts of collected data.
Furthermore, LLMs can be trained to recognize patterns and anomalies in communication data, crucial for preempting and resolving potential issues \cite{jin2023time,zhu2024llms}. For instance, if a UAV sends inconsistent data, LLMs could quickly detect anomalies and alert operators. LLMs can enhance scalability and adaptation in communication protocols, automatically learning and adapting to new protocols based on new data or operational changes, ensuring seamless communication. Pre-training LLMs with simulated data helps understand mission conditions and requirements, enabling real-time adaptation during missions for optimal performance.

This work is motivated by the potential of integrating LLMs into UAV communication systems. We comprehensively analyze existing LLM methods focusing on UAV integration to highlight advantages and limitations in expanding current UAV communication systems' capabilities. The review summarizes state-of-the-art LLM-integrated architectures, explores opportunities for LLM incorporation into UAV architecture, and addresses spectrum sensing and sharing concerning LLM integration. We aim to showcase how LLMs can optimize communication, adapt to new missions dynamically, and process complex data streams, enhancing UAV efficiency and versatility across various domains, including emergency response, environmental monitoring, urban planning, and satellite communication. Additionally, we address the legal, ethical, and technical challenges of deploying AI-driven UAVs, emphasizing responsible and effective integration, laying the groundwork for advancing UAV technology to meet future demands, and exploring innovative AI applications in UAV systems.

\subsection{Contributions}
The contributions of this paper are summarized as follows:

\begin{itemize}
    \item First, we present an in-depth analysis of various LLM architectures, assessing their suitability and potential for integration within UAV systems. This evaluation helps to understand the capabilities and effectiveness of state-of-the-art LLM models for different UAV applications.
    \item Then, we explore various LLM-based UAV architectures, providing a consolidated view of how UAV technologies evolve by integrating sophisticated AI models. Furthermore, areas that can further benefit from LLM integration are highlighted.
    \item After that, we discuss enhancing spectral sensing and sharing capabilities in UAVs through LLM integration. The impact of LLM integration on UAVs for optimized spectrum sensing, data processing, and decision-making is presented.
    \item Finally, we demonstrate how LLM integration with the UAV framework can extend the capabilities of UAVs in various sectors, including surveillance and reconnaissance, emergency response, delivery, and enhanced network connectivity during emergencies. Moreover, critical areas for future research essential for the successful and effective integration of LLMs with UAV systems are identified and highlighted.
\end{itemize}
\subsection{Related Surveys}
\begin{table*}[]
\caption{Summary of the existing surveys.}
\label{table-existing}
\resizebox{\textwidth}{!}{%
\begin{tabular}{|l|l|l|}
\hline
{\bf{Ref.}} & {\bf{Main Focus}} & {\bf{Key Findings}} \\ \hline
\cite{wang2024survey} & LLM-based autonomous agents & \begin{tabular}[c]{@{}l@{}}Comprehensive analysis of construction, application, and \\ evaluation of LLM-based agents in diverse fields.\end{tabular} \\ \hline
\cite{xi2023rise} & Development of LLM-based AI agents & \begin{tabular}[c]{@{}l@{}}Role of LLMs in artificial general intelligence and introduced\\ a novel framework based on brain, perception, and action \\ components to enhance the performance of LLM in complex \\ environments.\end{tabular} \\ \hline
\cite{wang2023aligning} & Aligning LLMs with human expectations & \begin{tabular}[c]{@{}l@{}}Focuses on the challenges such as misunderstanding, biased \\ outputs and factual inaccuracies in model-generated content. \\ It also provides an in-depth analysis of technologies that enhance \\ LLM alignments according to human expectations.\end{tabular} \\ \hline
\cite{zhu2023survey} & Deployment challenges of LLMs & \begin{tabular}[c]{@{}l@{}}Explores model compression techniques such as quantization, \\ pruning, and distillation to improve LLM efficiency.\end{tabular} \\ \hline
\cite{gao2024llm} & Challenges of LLM & \begin{tabular}[c]{@{}l@{}}Discusses the challenges of LLM related to dataset management, \\ tokenizer reliance, and pre-training costs.\end{tabular} \\ \hline
\cite{kaddour2023challenges} & Limitations of LLM & \begin{tabular}[c]{@{}l@{}}Focuses on operational challenges of LLM such as design flaws, \\ and inefficiencies of LLMs.\end{tabular} \\ \hline
\cite{sun2024generative} & GAI in UAV systems & \begin{tabular}[c]{@{}l@{}}Reviews GAI technologies for improving UAV communication, \\ networking, and security.\end{tabular} \\ \hline
\cite{liu2024generative} & GAI in UAV swarms & \begin{tabular}[c]{@{}l@{}}Surveys applications, challenges, and opportunities of GAI in \\ enhancing UAV swarm coordination and functionality.\end{tabular} \\ \hline
\multicolumn{1}{|l|}{\cite{bariah2024large}} &
  \multicolumn{1}{l|}{\begin{tabular}[c]{@{}l@{}}Large-GenAI models \\ for future wireless networks\end{tabular}} &
  \multicolumn{1}{l|}{\begin{tabular}[c]{@{}l@{}}Investigated the potential of Large-GenAI models \\ to enhance future wireless networks by improving wireless \\ sensing and transmission.\end{tabular}} \\ \hline
\multicolumn{1}{|l|}{\cite{du2023power}} &
  LLMs in wireless networks &
  \begin{tabular}[c]{@{}l@{}}Investigated the application of LLMs for developing advanced \\ signal-processing algorithms in wireless communication and \\ networking.\end{tabular} \\ \hline

\multicolumn{1}{|l|}{This paper} &
  \multicolumn{1}{l|}{LLM-integrated UAV systems} &
  \multicolumn{1}{l|}{\begin{tabular}[c]{@{}l@{}}A comprehensive analysis of various LLM architectures for UAV \\ systems, exploring their integration potential and impact across \\ different applications, including enhanced spectrum sensing and \\ sharing for autonomous decision-making. It also discusses how \\ LLM integration can extend UAV capabilities in various sectors, \\ including surveillance, emergency response, and delivery, while \\ identifying critical areas for future research to optimize their \\ integration and effectiveness.\end{tabular}} \\ \hline
\end{tabular}%
}
\end{table*}

The future holds promise for revolutionizing various domains; therefore, several recent review articles exist on the topic. For instance, \cite{gao2023retrieval,mialon2023augmented,schwartz2023enhancing}  investigate LLMs architectures, \cite{yin2023survey, zhang2023instruction, huang2022towards,liu2024large}  present the overview of the training process, fine-tuning, logical reasoning, and related challenges to address their limitations for broad adoption of LLM-based systems across domains.
In \cite{wang2024survey}, a comprehensive analysis is provided on LLM-based autonomous agents, focusing on their construction, application, and evaluation. These agents, equipped with sophisticated natural language understanding and generation capabilities, operate without human intervention. They interact with environments and users in complex ways, necessitating the integration of advanced AI techniques for tasks like communication and problem-solving across various domains such as social science, natural science, and engineering. Another work \cite{xi2023rise} delves into the development and use of LLM-based AI agents, emphasizing their role in advancing artificial general intelligence. LLMs are recognized as foundational for creating versatile AI agents due to their language capabilities, which are crucial for various autonomous tasks. The authors propose a framework based on brain, perception, and action components to enhance agents' performance in complex environments.
In \cite{wang2023aligning}, challenges and advancements in aligning LLMs with human expectations are critically examined. Concerns like misunderstanding instructions and biased outputs are addressed through technologies enhancing LLM alignment. Data collection strategies, training methodologies, and model evaluation techniques are explored to improve performance in understanding and generating human-like responses. Another study \cite{zhu2023survey} investigates challenges in deploying LLMs, particularly in resource-constraint settings. Model compression techniques like quantization, pruning, and knowledge distillation are discussed to improve efficiency and applicability.
While \cite{gao2024llm,kaddour2023challenges} investigate challenges of LLMs, including vast dataset management and high costs, they point out limitations that cannot be overcome merely by increasing model size.
\cite{sun2024generative} explores Generative Artificial Intelligence (GAI) applications in improving UAV communication, networking, and security performances. A GAI framework is introduced to advance UAV networking capabilities. \cite{liu2024generative} surveys the challenges of UAV swarms in dynamic environments, discussing various GAI techniques for enhancing coordination and functionality.
In \cite{bariah2024large}, the authors explore the potential of Large-GenAI models to enhance future wireless networks by improving wireless sensing and transmission. They highlight the benefits of these models, including enhanced efficiency, reduced training requirements, and improved network management. In another study \cite{ du2023power}, the authors investigated the application of LLMs for developing advanced signal-processing algorithms in wireless communication and networking. They explored the potential and challenges of using LLMs to generate hardware description language code for complex tasks, focusing on code refactoring, reuse, and validation through software-defined radios. This approach led to significant productivity improvements and reduced computational challenges.
Although \cite{sun2024generative,liu2024generative} focus broadly on GAI, the specific application of LLMs in UAV communication systems still needs to be explored. This gap highlights an area poised for investigation. Table \ref{table-existing} summarizes existing surveys' primary focus and critical findings.


\subsection{Organization}
The rest of the paper is organized as follows. In Section \ref{overview}, we present an overview of LLMs, introducing foundational concepts and developments in this area. Section \ref{LLM-UAV} is dedicated to exploring LLMs for UAVs, where we discuss the integration and adaptation of LLM technologies within UAV systems. Section \ref{architecture} focuses on network architectures for LLMs in UAV communications, examining the structural designs that support LLM functionalities in UAV networks. Section \ref{spectrum-management} addresses spectrum management and regulation for LLMs in UAVs. Section \ref{sec:application} explores applications and use cases of LLMs in UAV communications, outlining practical implementations and the benefits derived from these technologies. Section \ref{challenges} examines the challenges and considerations in implementing LLM-Integrated UAVs, discussing potential obstacles and operational considerations. Section \ref{future-directions} is dedicated to future directions and research opportunities, presenting potential areas for further exploration and development in the LLMs in UAVs. Finally, Section \ref{conclusion} concludes the paper with a summary of our findings and reflections on the broader implications of our research.

\section{Overview of Large Language Models (LLMs)} \label{overview}

LLMs undergo extensive data training, demanding high computational power and the integration of ML algorithms and deep neural network architectures to enable sophisticated language processing abilities \cite{jozefowicz2016exploring}. This training encompasses diverse and extensive datasets, including text from various sources such as books, websites, and articles, aiding the model in learning language structure, vocabulary, grammar, and contextual nuances \cite{jozefowicz2016exploring}. Integrating ML algorithms, LLMs interpret data, adjusting their learning process to language nuances and refining prediction and response generation \cite{zhang2023automl}. Typically based on transformer architecture, LLMs utilize self-attention mechanisms, simultaneously enabling parallel processing of every word in a sentence. This capability allows a comprehensive understanding of context, significantly enhancing the model's ability to manage long-range dependencies in text, thereby improving context awareness \cite{peng2024limitations}.
LLMs consist of multiple neural network layers, each performing complex computations. These layers process inputs sequentially, refining information progressively. With parameters ranging from millions to billions, they dictate how input data is transformed into outputs. During training, these parameters are adjusted to minimize prediction errors, progressively improving the model's ability to recognize complex patterns and relationships within the data \cite{naveed2023comprehensive}.

\begin{figure*}
  \centering
  \includegraphics[width=18cm]{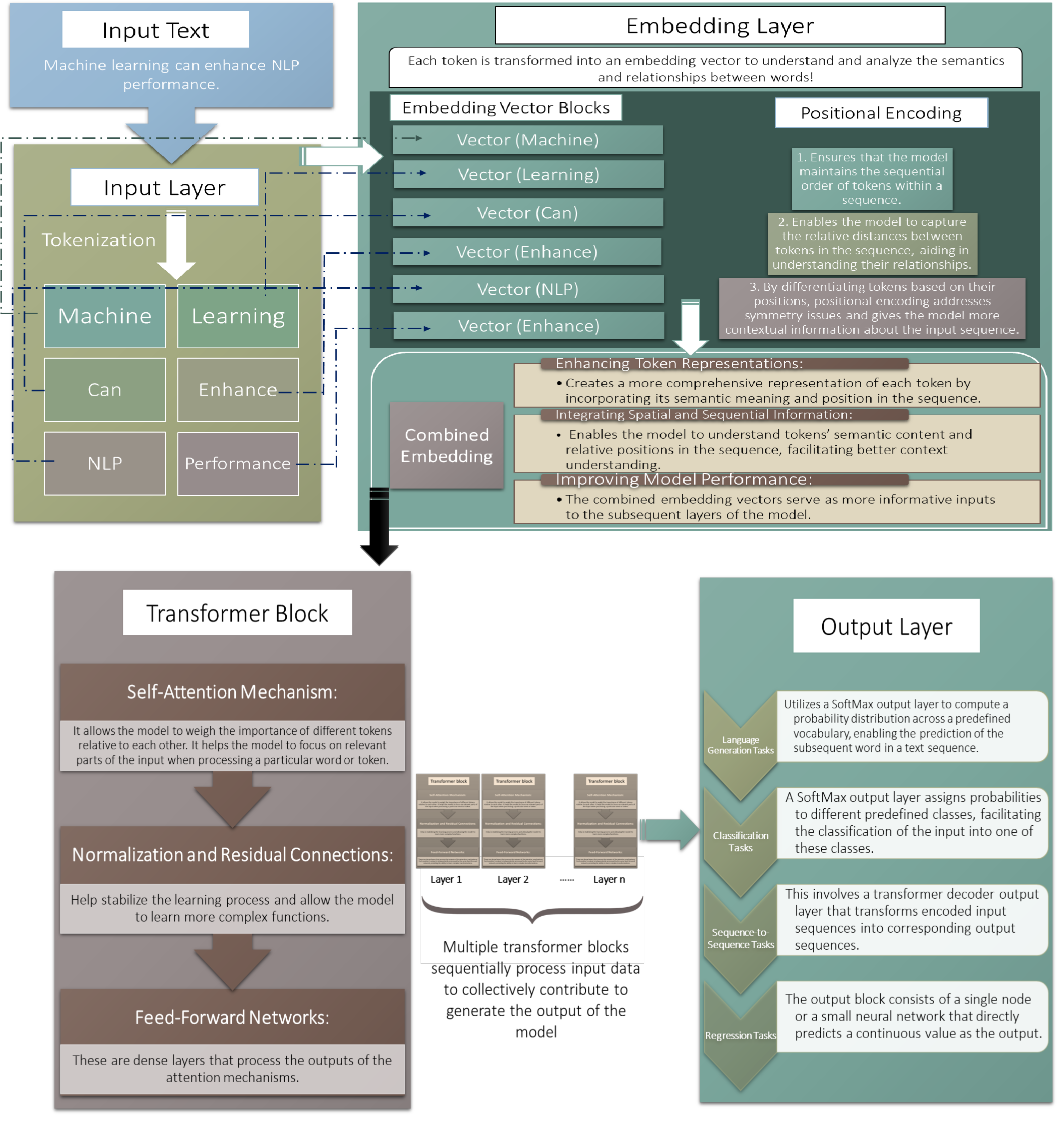}
  \caption{A general architecture of LLM incorporating input layer, embedding layer, transformer block, and output layer.}
  \label{fig:basicarc}
\end{figure*}

The specific architecture of LLMs, based on transformer architecture, incorporates attention mechanisms to weigh the importance of different parts of input data, essential for tasks like language understanding \cite{vaswani2017attention}. Positional encodings maintain word order, which is crucial for sequential natural language. Training involves adjusting parameters through gradient descent and backpropagation. Backpropagation calculates gradients of the loss function for each parameter, guiding parameter updates to enhance model performance. Gradient descent iteratively adjusts parameters to minimize prediction error. Through this process, LLMs continually learn from mistakes, generating coherent and contextually appropriate responses and improving their ability to handle complex language tasks. Fig. \ref{fig:basicarc}  illustrates the typical layers involved, showing how input text is transformed through multiple processing stages to produce output.

The broad landscape of LLMs in based on several essential models having unique characteristics tailored to specific linguistic tasks. For example, BERT \cite{devlin2018bert,alaparthi2020bidirectional} revolutionized the field by using a bidirectional training of transformers to improve context understanding, making it exceptionally practical for tasks such as question answering and language inference. Since its introduction, BERT has been widely adopted for improving search engine results, enabling more nuanced and context-aware responses to queries. Google has incorporated BERT into its search algorithms, significantly enhancing search accuracy by better understanding the intent behind users' queries \cite{sun2019bert4rec}. Moreover, Enhanced Representation through Knowledge Integration (ERNIE) \cite{sun2019ernie} extends BERT by incorporating structured knowledge, such as entity concepts, to enhance language comprehension. ERNIE is used for enhancing language understanding tasks in Chinese, showcasing superior performance in language inference and named entity recognition. It demonstrates the importance of integrating structured knowledge into pre-trained language models, especially for languages with complex structures like Chinese.

While the GPT series \cite{rosario2024generative,yenduri2024gpt} focuses on generating human-like text by predicting the next word in a sequence, showcasing remarkable proficiency in various generative tasks. The GPT models, particularly the later versions such as GPT-3, have been instrumental in creating advanced chatbots and virtual assistants. These models have been used to generate creative content, from writing articles to composing poetry, demonstrating their versatility in handling various innovative and conversational tasks.

On the other hand, T5 \cite{mastropaolo2021studying} adopts a unified framework that converts all text-based language problems into a text-to-text format, simplifying the process of training and applying transformers. T5 has been effectively used in summarization, translation, and classification tasks. Its flexible text-to-text approach can be fine-tuned for various language tasks without significantly changing the underlying model architecture \cite{rodriguez2022end}.

While eXtreme Learning NETwork (XLNet) introduces permutation-based training, which better captures the bidirectional context by predicting all tokens in a sequence rather than one at a time. XLNet's permutation-based training strategy allows it to outperform models like BERT on tasks requiring a deep understanding of context order, such as question answering and document ranking, making it particularly useful in academic and professional settings where the precise interpretation of information is crucial \cite{yang2019xlnet}.

Lastly, Bidirectional and Auto-Regressive Transformers (BART) combine an autoencoder and autoregressive approach, making it highly effective for sequence-to-sequence tasks and text generation. It has been used extensively in text summarization applications as it generates concise summaries of long documents without losing critical information. It is particularly useful in law and medicine files where extracting accurate information quickly from large texts is vital.
The next section details these models' working and highlights their opportunities for UAVs \cite{sankar2022comparative}.

Some less common LLM models have also been integrated into UAVs for innovative applications. In \cite{de2023semantic}, the authors employed UAVs to handle complex tasks involving semantically rich scene understanding by integrating LLMs and Visual Language Models (VLMs). The approach allows UAVs to provide zero-shot literary text descriptions of scenes, which are both instant and data-rich, suitable for applications ranging from the film industry to theme park experiences and advertising. The descriptions generated by this system achieve a high readability score, highlighting the system's ability to develop highly readable and detailed descriptions, demonstrating the practical use of microdrones in challenging environments for efficient and cost-effective scene interpretation.
The authors in \cite{ONEATA2021106943} integrated LLMs in UAV control by developing a multimodal evaluation dataset that combines text commands with associated utterances and relevant images, enhancing UAV command interpretation. The presented study evaluates the effectiveness of generic versus domain-specific speech recognition systems, adapted with varying data volumes, to optimize command accuracy. It also innovatively integrates visual information into the language model, using semi-automatic methods to link commands with images, providing a richer context for command execution.

In another work \cite{choutri2022multi}, the authors focused on enhancing human-drone interaction by implementing a multilingual speech recognition system to control UAVs. This work addresses the complexities and training challenges associated with traditional RF remote controls and ground control stations using natural, user-intuitive interfaces that recognize speech in English, Arabic, and Amazigh. The study developed a two-stage approach by initially creating a deep learning-based model for multilingual speech recognition, which was then implemented in a real-world setting using a quadrotor UAV. The model was trained on extensive records, including commands and unknown words mixed with background noise, to enhance its robustness for controlling drones across linguistic backgrounds.

In \cite{tagliabue2023real}, the authors explore the application of LLMs to enhance the resilience and adaptability of autonomous multirotors through REsilience and Adaptation (REAL) using the LLMs approach. REAL integrates LLMs into robots' mission planning and control frameworks, leveraging their extensive capabilities in long-term reasoning, natural language comprehension, and prior knowledge extraction. REAL utilizes LLMs to enhance robot resilience in facing novel or challenging scenarios, interpret natural language and log data for better mission planning, and adjust control inputs based on minimal user-provided information about robot dynamics. Experimental results demonstrate that this integration effectively reduces position-tracking errors under scenarios of controller parameter errors and unmodeled dynamics. Furthermore, it enables the multirotor to make decisions that avoid potentially dangerous situations not previously considered in the design phase, showcasing LLMs' practical benefits and potential in improving autonomous UAV operations.

\section {LLMs for UAVs} \label{LLM-UAV}
Due to the growing interest in LLM-integrated UAV systems across various applications, several recent work exists. For instance, in \cite{zhong2024safer}, the authors introduced a vision-based autonomous planning system for quadrotor UAVs to enhance safety. The system predicts trajectories of dynamic obstacles and generates safer flight paths using NanoDet for precise obstacle detection and Kalman Filtering for accurate motion estimation. Additionally, the system incorporates LLMs such as GPT-3 and ChatGPT to facilitate more intuitive human-UAV interactions. These LLMs enable Natural Language Processing (NLP), allowing users to control UAVs through simple language commands without requiring complex programming knowledge. They translate user instructions into executable code, enabling UAVs to execute tasks and provide feedback in natural language, simplifying the control process.  UAVs can operate in ad-hoc and mesh fashion to form dynamic networks without relying on pre-existing infrastructure. This makes them particularly valuable when establishing permanent network infrastructure is impractical, such as disaster response, military operations, or environmental monitoring. Both ad-hoc and mesh networks enhance UAVs' ability to configure and maintain connectivity as they move automatically. They continuously discover new neighbors and can adjust routes based on the network's topology and traffic conditions, improving scalability and flexibility \cite{jawhar2017communication}. Integrating LLMs into UAV communications enhances their ability to understand network conditions and generate insights based on the networks' characteristics, thereby highlighting their adaptability and responsiveness to quickly adapt to changing environmental conditions and operational demands. LLMs also help UAVs understand network traffic patterns for recommending adaptive protocols that reduce latency and increase throughput, particularly in the variable conditions common in these networks. They also assist in simulating or modeling the behavior of the networks under various scenarios, helping in planning and decision-making processes for UAV deployments. Therefore, the LLM incorporation can enhance data analysis, improving data exchange efficiency between UAVs.
LLMs, with their capability to process and learn from vast amounts of data, enable UAVs to make informed decisions about route planning, data forwarding, and network configuration. For instance, in response to UAV failures or environmental obstacles, LLMs can swiftly calculate alternative routes or reconfigure the network to sustain connectivity and performance. Moreover, LLMs foster higher levels of autonomy in UAVs by equipping them with advanced cognitive capabilities, allowing UAVs to understand and execute complex commands and interact more naturally with human operators or other autonomous systems.

Furthermore, LLMs can analyze data from UAVs (such as operational logs and telemetry data) to predict potential failures or maintenance needs before they occur. This predictive capability can significantly increase the reliability and lifespan of UAVs, reducing downtime and maintenance costs. Security is also a paramount concern in decentralized ad-hoc networks; LLMs can enhance security protocols by identifying potential threats through pattern recognition and anomaly detection and simulating attack scenarios to develop more robust security measures.
LLMs can also optimize allocating critical resources such as bandwidth and power within the UAV network. LLMs dynamically allocate resources by understanding and predicting network demands to maximize efficiency and extend UAV operational times. They improve the interface between human operators and the UAV network, providing more intuitive control and feedback systems, including generating natural language reports on network status or translating complex network data into actionable insights for decision-makers. In addition, LLMs address scalability challenges inherent in ad-hoc networks. They dynamically adjust network protocols and configurations as the number of UAVs changes, ensuring the network remains stable and efficient regardless of size. By integrating LLM capabilities, UAV ad-hoc networks can become more intelligent, responsive, and efficient, significantly enhancing their effectiveness across various applications.

This section provides a detailed overview of different LLMs and discusses the opportunities they bring for UAV-based communication systems.

\subsection{BERT-enabled UAVs}
As discussed in the previous section, BERT is an influential model in NLP, developed by researchers at Google and released in 2018 \cite{devlin2018bert}. BERT's development represented a turning point in NLP, providing a more nuanced and effective way for machines to process and understand human language by fully leveraging the context surrounding each word. BERT employed pre-training and fine-tuning stages. In pre-training, the model is trained on a large corpus of text with tasks designed to help it learn general language patterns. These tasks include predicting masked words in a sentence (i.e., Masked Language Model (MLM)) and predicting whether two sentences logically follow each other (i.e., Next Sentence Prediction (NSP)). After pre-training, BERT is fine-tuned with additional data tailored to specific tasks (i.e., question answering or sentiment analysis) \cite{koroteev2021bert,ravichandiran2021getting}.

The introduction of BERT significantly advanced the state of the art in a wide range of NLP tasks. It showed remarkable performance improvements on leaderboards for tasks such as named entity recognition \cite{ehrmann2023named,hakala2019biomedical}, sentiment analysis \cite{xu2019bert,alaparthi2021bert}, and especially question answering and natural language inference, where the full-sentence context from both directions can be crucial for understanding subtleties. In addition, BERT has inspired numerous variations and improvements, leading to the development of different models such as Robustly optimized BERT approach (RoBERTa) \cite{liu2019roberta}, Distilled Bidirectional Encoder Representations from Transformers (DistilBERT) \cite{sanh2019distilbert}, and A Lite BERT (ALBERT) \cite{lan2019albert} that use the original architecture and training procedures of BERT to optimize other factors such as training speed, model size, or enhanced performance.

Integrating BERT can significantly enhance UAV performance across various domains. For instance, in emergency response scenarios, BERT can help UAVs understand complex natural language commands from disaster management teams. Moreover, BERT can interpret and summarize information from UAV sensors and reports, making it particularly valuable in surveillance missions where quick summarization of extensive video data is essential. Moreover, BERT's ability to rapidly analyze and interpret data from multiple sources enables timely, informed decisions, crucial in environmental monitoring for assessing conditions like forest fires or pollution. Furthermore, its proficiency in parsing and understanding commands ensures precise coordination among multiple UAVs, which is critical for complex logistics operations involving supply delivery in challenging environments. Recently, in \cite{ luo2024language}, the authors introduced an innovative end-to-end Language Model-based fine-grained Address Resolution framework (LMAR) explicitly designed to enhance UAV delivery systems. Traditional address resolution systems rely primarily on user-provided Point of Interest (POI) information, often lacking the necessary precision for accurate deliveries. To address this, LMAR employs a language model to process and refine user-input text data, improving data handling and regularization with enhanced accuracy and efficiency in UAV deliveries. In another work \cite{silalahi2022named,silalahi2023transformer}, the authors develop enhanced security and forensic analysis protocols for UAVs to support increased drone usage across various sectors, including those vulnerable to criminal misuse. They introduce a named entity recognition system to extract information from drone flight logs. This system utilizes fine-tuned BERT and DistilBERT models with annotated data, significantly improving the identification of relevant entities crucial for forensic investigations of drone-related incidents. The authors in \cite{ fan2023research } focused on enhancing the target recognition capabilities of UAVs in intelligent warfare by constructing a standardized knowledge graph from large-scale, unstructured UAV data. The authors introduced a two-stage knowledge extraction model with an integrated BERT pre-trained language model to generate character feature encoding, which enhances the efficiency and accuracy of information extraction for future UAV systems.

\subsection{GPT-enabled UAVs}
GPT series developed by OpenAI represents a significant evolution in the design and capabilities of LLMs that enhance various natural language processing tasks such as text generation, translation, summarization, and question answering \cite{radford2018improving}. The first architecture, GPT-1, was introduced in June 2018, and it was based on the transformer model that uses a stack of decoder blocks from transformer architecture. GPT-1 was pre-trained on a language modeling task (predicting the next word in a sentence) using the BooksCorpus dataset, which comprises over 7,000 unique unpublished books (totaling around 800 million words). After this initial pre-training, supervised learning was fine-tuned for specific tasks \cite{ye2023comprehensive,radford2018improving}.

The GPT-2 was released in February 2019 and expanded significantly on its predecessor, featuring up to 48 layers in its largest version, with 1,600 hidden units, 48 attention heads, and 1.5 billion parameters \cite{radford2019language}. GPT-2 used a WebText dataset created by scraping web pages linked from Reddit posts with at least three upvotes. This resulted in a diverse dataset of around 40GB of text data. GPT-2 continued using the unsupervised learning approach, leveraging only language modeling for pre-training without task-specific fine-tuning. This demonstrated the model's ability to generalize from language understanding to specific tasks \cite{kalyan2023survey}. While GPT-3 released in June 2020 is one of the largest AI language models ever created, with 175 billion parameters. It includes 96 layers, with 12,288 hidden units and 96 attention heads \cite{brown2020language}. It was trained on an even more extensive and diverse dataset, including a mixture of licensed data, data created by human trainers, and publicly available data, significantly larger than GPT-2. GPT-2 and GPT-3 used an unsupervised learning model, demonstrating exceptional capabilities in learning from large datasets \cite{hendy2023good}.
GPT-4, built on an advanced transformer-style architecture, significantly expands in size and complexity compared to its predecessors, GPT-2 and GPT-3. This model has been fine-tuned using Reinforcement Learning from Human Feedback and employs publicly available internet data and data licensed from third-party providers. However, specific details related to the architecture, such as the model size, hardware specifications, computational resources used for training, dataset construction, and training methodology, have not been publicly disclosed \cite{achiam2023gpt}.

GPT series in UAVs represents an innovative intersection of AI and drone technology that can enhance UAVs' functionality, autonomy, and interaction capabilities in a broad range spanning from enhanced control systems to fully autonomous task execution \cite{tazir2023words,wang2017construction}. For example, GPT series integration allows UAVs to execute instructions provided in plain language with a high level of proficiency. For instance, the operator command to inspect the condition of a bridge at specific coordinates will instruct the UAV to devise a flight path and carry out all the necessary steps for bridge inspection without requiring manual inputs for each step. Similarly, it can generate detailed reports based on data collected during flights, and integration of these models with the UAV's sensors and data collection systems can automatically generate textual descriptions highlighting various aspects, such as mission outcomes and anomalies detected \cite{biswas2023prospective}. Accordingly, it can make it easier for human operators to understand what the UAV has observed without reviewing extensive raw data. For instance, Tazir et al. in \cite{ tazir2023words} integrated LLM system OpenAI's GPT-3.5-Turbo with UAV simulation systems (i.e., PX4/Gazebo simulator) to create a natural language-based drone control system. The system architecture is designed to allow seamless interaction between the user and the UAV simulator through a chatbot interface facilitated by a Python-based middleware. The Python middleware is the core component, establishing a communication channel between the chatbot (GPT-3.5-Turbo) and the PX4/Gazebo simulator. It processes natural language inputs from the user, forwards these inputs to the ChatGPT model using the OpenAI API, retrieves the generated responses, and converts them into commands that the simulator understands. ChatGPT provides guidance and support through PX4 commands and explanations, thus enhancing the interactivity and accessibility of UAV simulation systems. It also facilitates the control and management of UAVs through sophisticated AI-driven interfaces. In another work \cite{choudhury2024natural}, the authors integrated advanced GPT models and dense captioning technologies into autonomous UAVs to enhance their functionality in indoor inspection environments. The proposed system enables UAVs to understand and respond to natural language commands like humans, increasing their accessibility and ease of use for operators without advanced technical skills. The UAVs' dense captioning models facilitate this human-like interaction by analyzing images captured during flight to generate detailed object dictionaries. These dictionaries allow the UAVs to recognize and understand various elements within their environment, dynamically adapting their behavior in response to both expected and unexpected conditions, thus improving the efficiency and accuracy of UAVs for indoor inspections across various environmental conditions and applications.

Furthermore, in dynamic or complex environments requiring fast decisions, GPT series can assist by processing real-time data and communications, providing suggestions or automated decisions based on the data. For example, a search and rescue operation can analyze live video feeds and text reports from multiple UAVs, synthesize the information, and recommend areas to focus on or adjust search patterns \cite{de2023socratic}. It can also play a crucial role in enhanced collaborative UAV-to-UAV communication by establishing a decentralized swarm intelligence system where UAVs can share information and make group decisions. For instance, UAVs can use natural language to report their status and findings to each other, coordinate their actions based on shared goals, and optimize task distribution among the group without constant human intervention \cite{wang2023survey}. GPT series can also simulate various communication scenarios for UAV's training by generating realistic mission scenarios and responses, providing operators with robust training on handling different situations to enhance their response for real-world operations \cite{biswas2023prospective}.


\subsection{Text-to-Text Transfer Transformer (T5) for UAVs}
Google introduced the T5 model in October 2019 and adopted a novel and streamlined approach to handling various NLP tasks by reframing them as text-to-text problems \cite{raffel2020exploring}. Unlike traditional models that require different architectures for different tasks and produce varied outputs, T5 standardizes the input and output across all tasks \cite{ni2021sentence}. Each NLP task, such as translation, summarization, question answering, or text classification, is treated as generating new text from a given text. Consequently, T5 employs a single consistent model architecture for all tasks. This simplification streamlines both the model training and deployment pipeline, as the same model can be trained across multiple tasks with minimal architectural modifications \cite{mastropaolo2021studying}. For example, in a translation task where the input is English text, and the output is French text, both are treated merely as sequences of words. T5 is pre-trained on a large corpus of text in a self-supervised manner, primarily using a variation of the masked language model task, similar to BERT. This pre-training equips the model to understand and generate natural language effectively. Following this, T5 is fine-tuned on specific tasks by adjusting its training data to suit the text-to-text format. The versatility of T5 makes it suitable for a broad range of applications, including language translation, document summarization, and sentiment analysis, where it interprets text sentiment by producing descriptive tags. It also excels in question answering by generating appropriate textual answers \cite{48643}.

UAVs can integrate the T5 framework to enhance the efficiency of UAV operations. Similar to GPT and BERT, T5 also improves command interpretation and response generation for UAVs, where complex commands issued by operators in natural language are interpreted and converted into executable instructions for the UAV. T5 is also proficient in generating comprehensive mission reports based on the data collected by the UAV by summarizing key findings, highlighting anomalies, and describing the surveyed area for environmental monitoring or disaster response applications.
Moreover, T5 can perform real-time operations by processing data streams from UAV sensors and cameras, providing immediate, actionable insights. For instance, during a search and rescue operation, T5 could quickly summarize visual and sensor data to describe potential areas of interest or hazards, helping guide rescue efforts more effectively. At the same time, T5 can significantly enhance the performance of coordinated UAV missions by interpreting messages from one UAV and generating appropriate responses or commands for others, facilitating seamless teamwork for various applications ranging from managing flight patterns to avoiding collisions or coordinating timings for area surveillance.

T5-assisted UAV communication also enables automated troubleshooting and feedback, such as if a UAV encounters issues or anomalies during its operation, it can help by interpreting error messages or sensor data and generating troubleshooting steps or advice in natural language. This can also extend to providing real-time feedback to operators on mission progress or suggesting adjustments to improve operational efficiency. In addition, T5 can generate simulated mission scenarios and dialogues based on historical data or potential future situations for training purposes.

\subsection{eXtreme Learning NETwork (XLNet) for UAVs}
XLNet is an advanced NLP model developed jointly by researchers from Google and Carnegie Mellon University \cite{yang2019xlnet}. Unlike BERT, which employs an MLM approach (i.e., where some words in a sentence are randomly masked and predicted), XLNet uses a permutation-based training strategy. This approach considers all possible permutations of the words in a sentence during training, enabling the model to predict a target word based on all potential contexts provided by other words before and after it. This method significantly enhances the flexibility and depth of contextual understanding. In addition, the permutation-based training allows XLNet to capture a richer understanding of language context, unlike BERT, which focuses only on predicting masked words and might miss contextual nuances \cite{cortiz2021exploring,adoma2020comparative}.

Furthermore, XLNet avoids the discrepancies between pre-training and fine-tuning phases seen in BERT by not relying on word masking during training, leading to more consistent behavior across different operational phases. XLNet also merges strategies from autoregressive language modeling (e.g., GPT series) and autoencoding (e.g.,  BERT) by training autoregressively without adhering to a fixed sequence order. Instead, it predicts words based on varied permutations, enhancing its comprehension and generation capabilities \cite{rajapaksha2021bert}. Consequently, XLNet has demonstrated superior performance on several NLP tasks, including question answering, natural language inference, and document ranking, by effectively utilizing complete sentence structures for a deeper and more accurate contextual understanding \cite{li2020comparing,topal2021exploring,oneata2019kite}.

Integrating XLNet to UAVs can offer unique advantages due to its sophisticated approach to language processing \cite{li2024behavior,yao2024can}. XLNet's permutation-based training enables a more nuanced and comprehensive understanding of context, making it particularly effective for interpreting complex instructions or environmental data where the context can vary significantly. For instance, during search and rescue missions, where the operational environment is complex and dynamic, XLNet can provide more reliable interpretations of context-heavy commands in real time.
Similarly, since XLNet considers all permutations of input data, it can be more robust against noisy or incomplete inputs, which are common in real-world UAV tasks. This feature is especially beneficial in combat or disaster response scenarios where communications may be disrupted or incomplete. XLNet's ability to contextually predict missing information can maintain the effectiveness of UAV operations.

\subsection{Enhanced Representation through kNowledge Integration (ERNIE) for UAVs}
Baidu Research introduced ERNIE in June 2019 to integrate world knowledge into pre-trained language models \cite{sun2019ernie}. It represents a significant evolution in language understanding due to its novel approach of integrating structured world knowledge into the training of language models. Unlike conventional models that rely on vast amounts of textual data to learn language patterns, ERNIE enhances these models by incorporating knowledge graphs into the training process. Knowledge graphs are structured databases that store information about the world so that machines can understand and process by including entities (such as people, places, and things) and their relationships.

ERNIE is trained on both traditional text corpora and knowledge graphs. Including knowledge graphs allow ERNIE to understand and represent complex relationships and attributes associated with various entities \cite{sun2020ernie}. This training involves two key components: textual data and knowledge integration. Textual data is similar to other models like BERT or GPT, and ERNIE processes this vast amount of text to learn syntactic and semantic patterns of language. At the same time, the knowledge Integration component enables ERNIE to simultaneously learn from knowledge graphs, where it absorbs structured information about real-world entities and their interrelations. Thus, this process enables ERNIE to understand context from the linear text and a multi-dimensional perspective involving real-world facts and relationships. Integrating knowledge graphs offers ERNIE a deeper understanding of language semantics, as it can relate words and phrases to real-world entities and their properties. This capability allows it to perform better on tasks that require nuanced understanding, such as question answering and named entity recognition \cite{sun2021ernie,xiao2020ernie}.

In addition, ERNIE’s ability to draw upon external knowledge helps it provide contextually appropriate responses or analysis, especially when background knowledge about specific topics is crucial. It can also better handle ambiguity in language, as the additional data from knowledge graphs provides clarity on potentially confusing or unclear text based on the broader context of the entities involved \cite{yu2021ernie}. The applications of ERNIE are broad and impactful, particularly in areas where deeper understanding and contextual awareness are necessary. For example, ERNIE can leverage its integrated knowledge base to answer complex questions that require understanding beyond the text, such as historical facts or specific details about people or places. It also improves the performance of semantic search engines by understanding the deeper meanings of queries regarding the knowledge it has learned, offering more relevant and precise answers.

 ERNIE's unique capability to integrate structured world knowledge from knowledge graphs with textual data can substantially benefit UAV communication. For example, ERNIE can interpret complex, context-dependent commands operators issue more effectively than traditional language models. For instance, if an operator gives a command involving geographic or operational terms, ERNIE's integration of knowledge graphs allows it to understand and execute the command more accurately. This is crucial during complex missions in unfamiliar territories where a precise understanding of local geography and terms is necessary. ERNIE also demonstrates effective autonomous decision-making capabilities based on environmental data and mission objectives, as it can process both the current mission data and integrated knowledge to make informed decisions. For example, in environmental monitoring, ERNIE could identify specific features or anomalies in the landscape based on its broader understanding of environmental science, aiding in more effective data collection and analysis.

ERNIE also exhibits high real-time situational awareness attributes during critical missions such as search and rescue or disaster response, where ERNIE can apply its semantic understanding to interpret real-time data inputs (e.g., visual or sensor data) against its knowledge graph. This helps identify relevant entities or situations quickly, such as recognizing areas historically known to be hazardous or interpreting signs of human activities in remote sensing data. In multiple UAV scenarios, ERNIE can facilitate better communication and coordination by understanding and managing the information exchange between UAVs. It can interpret and prioritize communication-based on the relevance and urgency related to the mission's objectives, using its semantic understanding to ensure that UAVs operate harmoniously.

Furthermore, in the context of training purposes, ERNIE can generate context-rich simulation scenarios that incorporate real-world knowledge into training exercises that assist in developing a better understanding of how to interact with UAVs in complex scenarios, enhancing their preparedness for real-world operations. Similarly to other LLMs, after mission completion, ERNIE can assist in generating detailed incident reports and debriefings that include observational data and contextual insights based on the integrated knowledge to provide a semantic analysis of the mission outcomes. Accordingly, by leveraging its ability to integrate and utilize extensive knowledge graphs alongside textual data, ERNIE can significantly enhance the capabilities of UAV communication systems, making them more intelligent, responsive, and effective in complex operational environments. This makes ERNIE particularly valuable for advanced UAV applications where conventional language models might fail to understand and process complex contextual information.

\subsection{Bidirectional and Auto-Regressive Transformers (BART) for UAVs}
Facebook developed BART that combines the strengths of both auto-encoding and auto-regressive techniques within the transformer framework, making it exceptionally effective for sequence-to-sequence tasks \cite{lewis2019bart}. Unlike BERT, which is primarily designed for understanding and predicting elements within the same input text, BART is optimized for tasks requiring text generation or transformation. It is trained by corrupting text with various noising functions, such as token masking, text infilling, and learning to reconstruct the original text \cite{bhattacharjee2020bert,koroteev2021bert}. The BART training equips it to handle a wide range of applications, including text summarization, where it can generate concise versions of longer documents, and text generation, suitable for creating content or generating dialogue. In addition, BART's capabilities extend to machine translation and data augmentation, making it a versatile tool for transforming input text into coherent and contextually appropriate output sequences \cite{van2011cognitive}.

BART integration into UAVs offers several advantages, particularly in tasks that involve complex text processing and generation. For example, BART can enhance the formulation and interpretation of mission reports, automatically generating concise summaries from extensive surveillance data or sensor readings, thus aiding in quicker decision-making and briefing. BART is also proficient in generating coherent text sequences for automated responses or instructions to UAV operators, specifically in scenarios requiring fast and accurate communication.

Furthermore, BART can improve real-time strategy adjustments during search and rescue operations to interpret incoming data and provide updated mission objectives or directions based on the evolving scenario. It also can transform noisy, incomplete textual data into intelligible information,  making it particularly valuable in dynamic and challenging environments for UAV operations to ensure that communications remain clear and contextually relevant despite the complexities involved.

\subsection{Comparison of different LLMs for UAVs}
The comparison of different LLMs (i.e., BERT, GPT, T5, XLNet, ERNIE, and BART) for UAVs reveals distinct capabilities tailored to various aspects of UAV operations, reflecting their unique architectures and training approaches. For instance, BERT excels at understanding context from both directions around a word, making it highly effective for interpreting complex commands and extracting relevant information from mission data. It is particularly suited for tasks where a precise understanding of sensor data or operational directives is critical, such as in surveillance or monitoring missions where deep contextual knowledge is crucial. In contrast, GPT specializes in generating coherent and extended text outputs, which are beneficial for creating detailed mission reports or conducting dialogues. This model is ideal for UAV training simulations that require narrative-style updates or interactive communications for generating operational logs or debriefing reports.

Whereas T5 exhibits high versatility and converts any text-based task into a text-to-text format, simplifying the processing of diverse types of communication. It proves effective in UAV communication tasks such as translating communications between different languages or protocols, summarizing extensive exploration data, and transforming raw sensor outputs into actionable text formats. On the other hand, XLNet employs permutation-based training and understands language context more flexibly and comprehensively than BERT. This model is helpful in complex, dynamic operational environments such as search, rescue, and disaster response, where interpreting and responding to context-heavy instructions in real-time is essential.

Similarly, ERNIE enhances the semantic understanding of language by integrating external knowledge through knowledge graphs, making it well-suited for missions that require a deep understanding of specific terminologies or concepts, such as environmental monitoring applications that involve specific ecological data. While BART compromises the benefits of auto-encoding and auto-regressive models, it excels at understanding and generating text. It is ideal for developing precise and contextually accurate mission reports or instructions for summarizing detailed surveillance data, where maintaining the integrity of information and its concise presentation are crucial.

Therefore, in conclusion, BERT and XLNet are highly effective at understanding commands due to their profound contextual understanding, with XLNet providing additional flexibility in dynamic contexts. Meanwhile, GPT and BART excel at creating coherent, extensive texts, with BART offering additional capabilities in text transformation tasks. T5 offers broad applicability across text transformation tasks, making it versatile for various communication needs. ERNIE stands out in scenarios where integrating specialized knowledge is essential for accurate operation and decision-making. Accordingly, each model can be incorporated based on the UAV mission's specific requirements to ensure that communication remains effective and efficient, tailored to the complexities and challenges of UAV operations. Table \ref{table-LLM} highlights various LLM models, including their key features, applications in the UAV domain, and challenges for integration in UAV systems.

\begin{table*}[]
\label{table-LLM}
\caption{Overview of different state-of-the-art LLM models for UAVs.}
\renewcommand{\arraystretch}{1.5}
\LARGE
\fontsize{10pt}{12pt}\selectfont
\resizebox{\textwidth}{!}{%
\begin{tabular}{|l|l|l|l|}
\hline
\textbf{LLM Model} &
  \textbf{Key Features} &
  \textbf{Applications in UAV domain} &
  \textbf{Challenges} \\ \hline
\begin{tabular}[c]{@{}l@{}}BERT \\ (2018, Google)\\ \cite{devlin2018bert}\end{tabular} &
  \begin{tabular}[c]{@{}l@{}}1. Contextual Language Understanding: Consider the \\ context from both the left and the right side of a word \\ within a sentence.\\ 2. Pre-training and Fine-Tuning Approach: Initially, \\ BERT is pre-trained on a large text corpus using tasks \\ such as MLM and NSP. This pre-training helps it understand \\ general language patterns. It is then fine-tuned with task-\\ specific data, enhancing its performance on particular NLP\\ tasks. \\ 3. Transferability Across Different Tasks: Once pre-trained, \\ BERT can be fine-tuned for a variety of NLP tasks\end{tabular} &
  \begin{tabular}[c]{@{}l@{}}Automated report generation, sentiment analysis of \\ ground communications,  language translation for \\ international operations, anomaly detection in textual \\ data, predictive maintenance from log analysis,  real-time \\ decision support Systems,  multi-UAV coordination through\\ NLP, training simulators with natural language interaction\end{tabular} &
  \multirow{6}{*}{\begin{tabular}[c]{@{}l@{}}1. Computational Requirements: All of these models are \\ computationally intensive, requiring significant processing \\ power, memory, and energy resources, limited in UAV\\ platforms due to their need for lightweight and efficient \\ designs. \\ 2. Real-Time Processing: The complexity of these models \\ often leads to slower processing times, making it difficult to \\ meet the real-time data processing demands of UAV \\ operations, where decisions need to be made quickly \\ and on-the-fly. \\ 3. Model Integration: Seamlessly integrating these LLMs \\ into existing UAV systems is complex, as it involves \\ compatibility with current hardware and software and \\ ensures that the models can interact effectively with \\ other onboard systems like sensors and navigational aids. \\ 3. Data Security and Privacy: Given that UAVs often collect \\ and process sensitive data, ensuring the security and privacy \\ of the data when using LLMs is crucial. This includes \\ protecting data from unauthorized access and ensuring \\ compliance with relevant privacy regulations. \\ 4. Latency in Communication Links: For models that require \\ more processing power than what's available on the UAV, \\ offloading data to cloud servers for processing can introduce \\ communication latency, which can adversely affect mission-\\ critical operations. \\ 5. Model Maintenance and Updation: LLMs need continual \\ updates and maintenance to perform optimally, which can \\ be challenging to manage, especially in deploying these \\ models in dynamic operational environments. \\ 6. Training Data and Model Robustness: Ensuring that these \\ models are trained on relevant, high-quality data and can \\ handle the diversity of scenarios encountered in UAV \\ missions without failing or producing erroneous outputs \\ is critical.\end{tabular}} \\ \cline{1-3}
\begin{tabular}[c]{@{}l@{}}GPT\\ 2018-2020\\ \cite{ye2023comprehensive}\end{tabular} &
  \begin{tabular}[c]{@{}l@{}}1. Text Generation Capabilities: GPT excels at generating \\ coherent, contextually appropriate text, making it ideal \\ for creating detailed mission reports and operational updates. \\ 2. Language Understanding and Translation: GPT can process \\ and translate multiple languages, facilitating international \\ operations and communication between different \\ language-speaking teams. \\ 3. Scalability Across Various Tasks: The model can adapt \\ to different NLP tasks without significant modifications, \\ allowing for seamless integration into diverse UAV \\ operational frameworks.\end{tabular} &
  \begin{tabular}[c]{@{}l@{}}Automated instruction execution, Mission report generation,\\ Real-time data analysis and recommendations, Dialogue systems \\ for operator interaction, Enhanced autonomous decision-making,\\ Language translation for multinational coordination, Simulated \\ training scenarios\end{tabular} &
   \\ \cline{1-3}
\begin{tabular}[c]{@{}l@{}}T5\\ 2019, Google\\ \cite{mastropaolo2021studying}\end{tabular} &
  \begin{tabular}[c]{@{}l@{}}1. Unified Text-to-Text Framework: T5 simplifies handling \\ various NLP tasks by converting all tasks into a uniform \\ text-to-text format, making it adaptable for different \\ communication needs. \\ 2. Comprehensive Pre-training: T5 is pre-trained on a \\ diverse corpus in a self-supervised manner, equipping \\ it with a broad understanding of language that can be\\ fine-tuned for specific UAV tasks. \\ 3. Scalable Across Tasks: The versatility of T5 allows it \\ to handle anything from translation to summarization, \\ making it highly applicable for multifaceted UAV \\ operations.\end{tabular} &
  \begin{tabular}[c]{@{}l@{}}Command interpretation and execution, Real-time data \\ summarization, Automated mission reporting, Translation and \\ multilingual communication, Interactive training simulations,\\ Operational decision support, Enhanced data analysis and \\ insight generation\end{tabular} &
   \\ \cline{1-3}
\begin{tabular}[c]{@{}l@{}}XLNet\\ 2019, Google and \\ CMU\\ \cite{yang2019xlnet}\end{tabular} &
  \begin{tabular}[c]{@{}l@{}}1. Permutation-Based Training: It involves permuting \\ the input tokens in all possible orders, allowing it to \\ understand the context more comprehensively and \\ flexibly than traditional models. \\ 2. Robust Contextual Understanding: By capturing \\ richer context from both previous and subsequent \\ tokens in the input, XLNet offers superior performance \\ in understanding complex instructions and contextual \\ nuances. \\ 3. Adaptability to Incomplete Inputs: XLNet is especially\\ effective at handling noisy or incomplete data, making it \\ robust in unpredictable environments.\end{tabular} &
  \begin{tabular}[c]{@{}l@{}}Complex command interpretation, Enhanced data analysis \\ for surveillance, Real-time decision support, Mission debriefing \\ and report generation, Robust communication in noisy \\ environments, Simulated training and scenario modeling, \\ Multi-UAV coordination and strategy development\end{tabular} &
   \\ \cline{1-3}
\begin{tabular}[c]{@{}l@{}}ERNIE\\ 2019, Baidu\\ \cite{sun2019ernie}\end{tabular} &
  \begin{tabular}[c]{@{}l@{}}1. Integration of Structured Knowledge: ERNIE incorporates\\ structured world knowledge from knowledge graphs, \\ enhancing its understanding of complex, context-dependent \\ language. \\ 2. Enhanced Semantic Comprehension: By leveraging \\ knowledge graphs, ERNIE can interpret and respond to \\ commands that involve specific terminology or require \\ deep contextual understanding. \\ 3. Improved Decision Making: ERNIE's ability to draw \\ from a vast array of structured information enables it \\ to make more informed decisions based on comprehensive\\ environmental data.\end{tabular} &
  \begin{tabular}[c]{@{}l@{}}Context-dependent command interpretation, Autonomous \\ decision-making, Enhanced situational awareness, Multi-UAV \\ coordination, Real-time data analysis and summarization, \\ Simulation and training enhancement, Incident reporting and \\ debriefing\end{tabular} &
   \\ \cline{1-3}
\begin{tabular}[c]{@{}l@{}}BART\\ 2019, Facebook\\ \cite{lewis2019bart}\end{tabular} &
  \begin{tabular}[c]{@{}l@{}}1. Effective Text Generation and Transformation: BART excels \\ in generating coherent text and transforming input data into \\ structured outputs, ideal for creating detailed mission reports\\ and communication. \\ 2. Robustness in Text Infilling and Correction: It can effectively \\ handle and correct corrupted text, making it valuable in scenarios \\ where data may be incomplete or noisy. \\ 3. Versatility in NLP Tasks: BART's architecture combines both \\ encoder and decoder components, allowing it to perform a wide \\ range of NLP tasks efficiently.\end{tabular} &
  \begin{tabular}[c]{@{}l@{}}Automated mission reporting, Real-time communication \\ enhancement, Data infilling and error correction, Multi-UAV \\ coordination commands, Operational brief generation, Interactive \\ training systems, Decision-making support in dynamic scenarios\end{tabular} &
   \\ \hline
\end{tabular}%
}
\end{table*}

\section{Network Architectures for Integrating LLMs in UAVs} \label{architecture}

\begin{figure*}
  \centering
  \includegraphics[width=16cm]{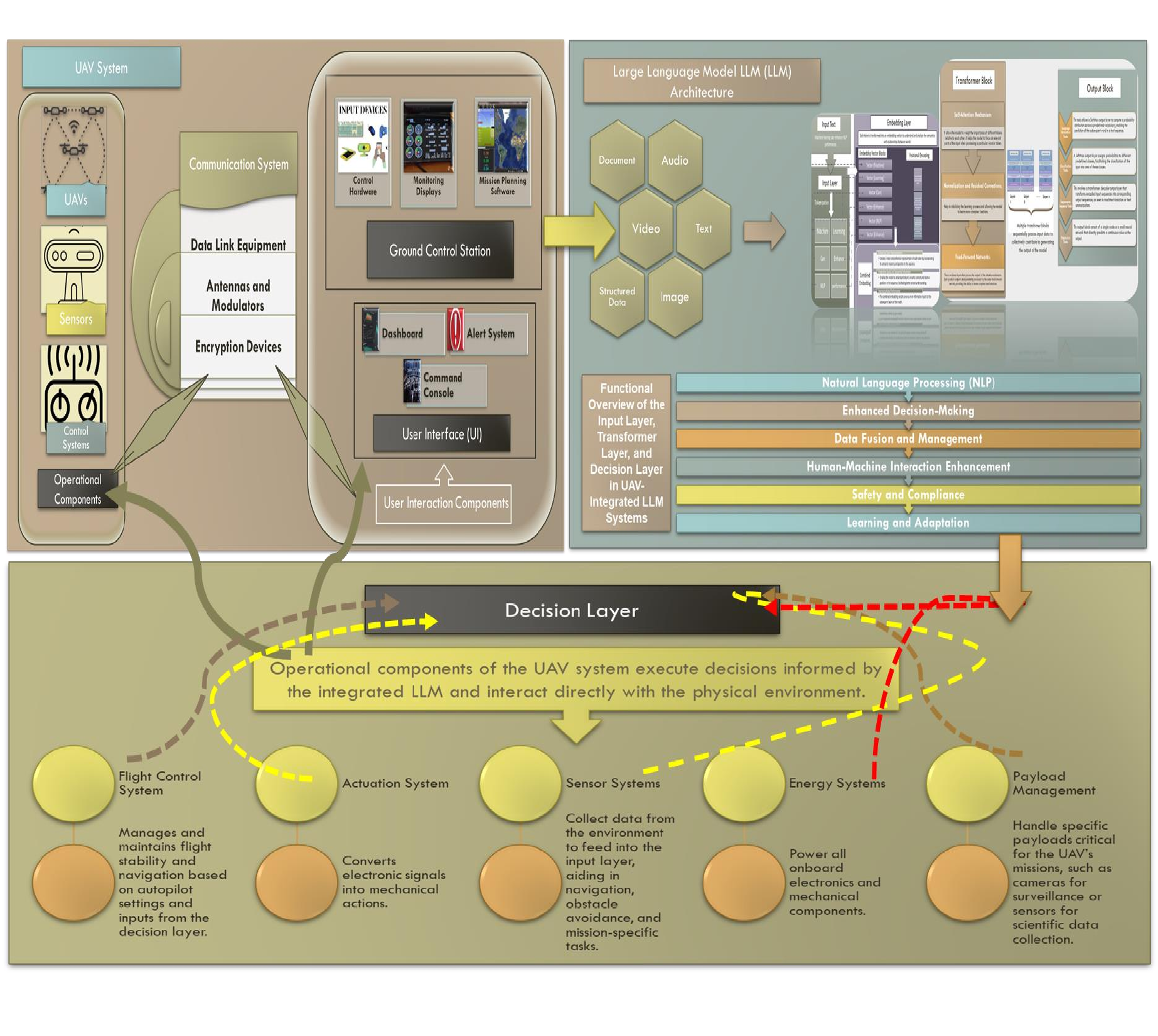}
  \caption{Comprehensive architecture of LLM-integrated UAV systems.}
  \label{fig:uavarc}
\end{figure*}

Integrating LLMs with UAVs involves deploying advanced language processing capabilities to enable sophisticated decision-making and interaction abilities. The UAV platform consists of essential hardware, including the UAV itself equipped with flight control hardware, sensors like cameras and LiDAR, and communication modules such as Wi-Fi, LTE, and satellite. It also includes small-scale onboard computers for real-time data processing. In LLM integration, lightweight versions of LLMs are directly deployed on UAVs for rapid and autonomous decision-making through Edge AI. For more complex computations, UAV data is transmitted to cloud servers where more robust LLMs perform analyses, and the results are then sent back to the UAV. Ground control stations support these operations, allowing operators to monitor and control UAVs remotely via direct line-of-sight or satellite communication, using secure data links for data transmission.
The operation of this system involves several key functions. UAVs collect data via onboard sensors, capturing visual imagery, environmental data, or specific readings relevant to their missions. This data is either processed locally or sent to a ground station or cloud server, depending on the task's complexity and onboard processing unit capabilities. The embedded LLM on the UAV processes data for simple tasks to make real-time decisions. For more complex decisions, data is sent to the cloud, where powerful LLMs analyze it, make decisions, or generate insights, which are then transmitted back to the UAV. Based on this processed data and decisions made by the LLMs, the UAV executes actions such as optimizing its flight path, interacting with the environment, or performing specific tasks like delivery, surveillance, or data collection.
Feedback and learning are integral to this system, where data from missions are used to retrain or refine LLMs, improving their accuracy and decision-making capabilities. This continuous feedback loop helps the model adapt to specific environments for optimal task performance. The integration of LLMs with UAVs thus offers significant enhancements to UAV operations, opening up vast possibilities for improved capabilities and effectiveness.
Fig. \ref{fig:uavarc} illustrates the comprehensive architecture of a UAV system integrated with an LLM, where the UAV collects data from sensors. This data, encompassing various types such as text, audio, and video, is input into the integrated LLM architecture. The LLM processes this data and outputs results to the decision layer, which then issues commands to operational components, including the flight controller, sensor systems, energy systems, and payload management systems.

Besides, ground control and base stations are key elements in the UAV operations infrastructure, serving as command and control centers that handle everything from flight authorization and monitoring to data processing and deployment management. Integrating LLMs with ground control and base stations significantly enhances UAV management and operation. For example, LLMs can significantly improve communication between UAVs and their control stations by interpreting and processing natural language commands or queries. It allows operators to interact with UAVs more intuitively, making complex commands simpler to execute and reducing the potential for human error.

LLMs can process real-time data received from UAVs at the ground control stations to make instant decisions regarding flight paths, mission adjustments, and responses to changing environmental conditions. LLMs can also analyze vast quantities of data much faster than humans, providing critical insights that enable quick decision-making to optimize UAV operations and ensure mission success. In addition, LLMs can utilize historical and real-time data to predict potential issues before they arise, such as mechanical failures, battery depletion, or adverse weather conditions. This predictive capability ensures that preventative measures can be taken in advance, enhancing the safety and reliability of UAV operations.

Furthermore, LLMs can automate routine tasks such as flight scheduling, monitoring UAV conditions, and managing data collection to improve efficiency for complex decision-making and operational strategy. LLMs can also signficantly contribute to improved data handling and analyzing by automatically categorizing data, extracting relevant information, and generating comprehensive reports. Moreover, they can analyze imagery and sensor data to identify patterns or anomalies, aiding in missions such as surveillance, environmental monitoring, and infrastructure inspection. LLMs can create detailed simulations and training scenarios based on accumulated data, providing operators with realistic and varied training experiences and improving the skills of UAV operators to ensure they are better prepared for complex operational scenarios.

Moreover, with their advanced pattern recognition capabilities, LLMs integrated into ground and base stations can enhance security protocols. They can detect potential cyber threats and unauthorized access attempts, ensuring UAV operations are protected against digital intrusions. In addition, LLMs can optimize resource allocation by predicting the best use of available UAVs and support equipment based on mission requirements. LLMs can also facilitate better interoperability between systems and software used in UAV operations to ensure seamless integration and communication across diverse platforms by acting as a bridge that understands and translates between various data formats and protocols. Thus, leading to efficient management, superior decision-making support, enhanced safety, and improved effectiveness of UAV missions. This broad application of LLMs lays the groundwork for their targeted use in enhancing spectrum sensing capabilities.

Moreover, given the pivotal role of spectrum sensing in ensuring effective Radio Frequency (RF) communication for UAVs, especially in complex or congested environments, the integration of LLMs proves immensely beneficial and can significantly enhance UAVs' spectrum sensing capabilities through sophisticated data processing techniques. This integration deepens the understanding of dynamic RF conditions, which are prevalent in areas with shared frequencies or high interference levels and enables UAV systems to identify and utilize optimal frequency bands intelligently. Such capabilities drastically improve the reliability and efficiency of their communication networks, which are crucial for maintaining robust links and ensuring the successful execution of UAV operations in RF-dense environments where traditional methods might fail. Consequently, this survey highlights the critical need for LLM integration in spectrum sensing and thoroughly explores its opportunities and challenges in the subsequent section.

\section{Spectrum Management and Regulation in LLMs-assisted UAVs} \label{spectrum-management}
UAVs depend on RF communication for various tasks, including remote control, telemetry, data transmission, and connectivity with ground stations. Spectrum sensing is a key technology that enhances UAVs' RF communication capabilities by enabling them to identify and utilize appropriate frequency ranges crucial for their missions. Moreover, it is particularly critical in environments where UAVs share frequency bands or encounter rapidly changing RF conditions \cite{wei2020performance, jasim2021survey}. Accordingly, by accurately sensing the spectrum, UAVs can dynamically adjust their communication parameters, such as channel selection and power control, to prevent interference with primary users and optimize their communication performance \cite{wang2018enabling}. In addition, spectrum sensing enhances the operational efficiency of UAVs by enabling them to make informed decisions about frequency band selection, thereby ensuring efficient utilization of available spectrum resources and minimizing the risk of interference with existing wireless systems \cite{shamsoshoara2019distributed, shang2020spectrum}.

Furthermore, spectrum sensing also plays an essential role in enabling cognitive radio capabilities \cite{zhang2016spectrum, wei2020performance}, dynamic spectrum access \cite{lin2020dynamic}, interference avoidance \cite{zhang2016spectrum}, and ensuring regulatory compliance for UAV communication systems \cite{massaro2017next}. For example, cognitive radio allows UAV systems to adaptively select and switch between different frequency channels or bands based on real-time spectrum sensing results, enabling UAVs to locate and utilize the most suitable, least congested, and interference-free frequency bands for reliable and efficient communication \cite{santana2018cognitive}. Dynamic spectrum access allows UAVs to access available spectrum resources, dynamically ensuring that UAVs can avoid interference with existing users while optimizing their communication links. In addition, spectrum sensing facilitates coexistence and interference avoidance by enhancing UAVs' capabilities in detecting the presence of other RF devices or systems in their vicinity. If interference or potential conflicts are detected, UAVs can autonomously or semi-autonomously change their operating frequency or adjust their communication protocols to avoid interference \cite{zhang2018spectrum}.

\subsection{Regulatory Frameworks and Compliance Considerations}
Regulatory bodies worldwide, such as the Federal Communications Commission (FCC) in the United States, establish guidelines for spectrum use to ensure fair access and prevent conflicts among various technologies and services, including UAVs. These guidelines designate specific frequency bands for UAV use to avoid conflicts with commercial, residential, and emergency communications to balance the growing demand for UAV services with the needs of traditional spectrum users. These authorities have established rules for dynamic spectrum access, particularly in bands where UAVs share the spectrum with other devices. This framework involves protocols and technologies that enable UAVs to detect and utilize vacant frequencies without interfering with incumbent users. Compliance with these frameworks is essential for legal and efficient UAV operations.

To ensure compliance, UAV operators must consider several key aspects. For instance, UAVs must be equipped with advanced spectrum sensing technologies that reliably identify available and occupied channels to prevent unauthorized use of occupied frequencies. UAVs must also operate to minimize interference with other spectrum users, adhering to power limits, frequency boundaries, and operational protocols designed to mitigate the risk of signal interference \cite{bayhan2019smart}. Moreover, it is necessary to implement software solutions that help manage spectrum usage and ensure adherence to local and international regulations to automate many aspects of spectrum management, reducing the burden on UAV operators and decreasing the risk of non-compliance.

\subsection{Integrating LLMs in UAVs Spectrum Management}
Recent research has significantly advanced the application of spectrum sensing and sharing in UAV operations, focusing on several key directions to enhance communication efficiency and mitigate interference. Shen et al. \cite{shen2019uav} introduced a 3D spatial-temporal sensing approach that leverages UAV mobility for dynamic spectrum opportunity detection in heterogeneous environments. The authors in \cite{liu2018spectrum} and  \cite{huang2018cognitive, huang2019cognitive} developed methods to optimize spectrum sensing and sharing in cognitive radio systems, improving UAV communication performance by managing interference with ground links. Chen et al. \cite{chen2019interference} focused on spectrum access management among UAV clusters to reduce interference, while Xu et al. \cite{xu2018efficient} focused on transmit power allocation and trajectory planning in UAV relay systems for effective data relay between devices.

In another work \cite{qiu2019blockchain}, Qiu et al. utilized blockchain technology to ensure privacy and efficiency in spectrum transactions between terrestrial and aerial systems. Hu et al. \cite{hu2018uav} focused on the strategies for spectrum allocation using contract theory to balance the interests of macro base stations and UAV operators. Azari et al. \cite{azari2020uav} compared underlay and overlay spectrum sharing mechanisms in densely populated urban scenarios, emphasizing the effectiveness of overlay strategies for maintaining service quality for both UAVs and ground users.
While significant advancements have been made in spectrum sensing and sharing technologies for UAVs, the integration of LLMs has not been widely explored in the existing research. Integrating LLMs can revolutionize the UAV domain by enhancing spectrum sensing capabilities, enabling more dynamic and efficient use of communication frequencies \cite{ghasemi2024accelerating}. LLMs can interpret and analyze the vast amounts of data generated by spectrum sensors on UAVs. With their advanced natural language processing capabilities, they can extract meaningful insights from unstructured data, facilitating intelligent decision-making in real time. LLMs can also predict spectrum availability and potential interference by analyzing historical data and current communication patterns. Thus, UAVs can proactively adjust their communication parameters, such as channel selection and power levels, to maintain optimal performance.

Furthermore, LLMs can process sensor data and identify patterns indicating potential frequency conflicts or areas of congestion. UAVs can then autonomously make adjustments to avoid these issues, enhancing operational efficiency and reducing the risk of communication failures. In addition, LLMs can contribute to cognitive radio enhancements by assisting in making more informed choices about frequency selection by providing a deeper analysis of spectrum conditions and user behavior. This integration enhances UAVs' ability to select the least congested and most efficient channels. LLM's continuous learning and adaptation ability can also optimize UAVs' spectrum access strategies, ensuring they utilize the best available frequencies based on real-time data and sophisticated algorithms. LLMs can also ensure that UAVs operate within the legal spectrum allocations by continuously monitoring compliance parameters and adapting to regulatory changes. LLMs can also contribute significantly to interference management and adherence to regulatory frameworks by analyzing communication patterns and environmental data. LLMs can detect potential interference sources more accurately and suggest immediate corrective actions to avoid them.

\section{Applications and Use Cases of LLMs in UAVs} \label {sec:application}
\subsection{Surveillance and Reconnaissance Applications}
LLMs offer advanced cognitive and analytical abilities that can significantly enhance the efficiency, accuracy, and effectiveness of UAV surveillance systems \cite{thakur2021artificial}. With LLM integration, UAVs can process and analyze large volumes of visual data more efficiently, enabling real-time image recognition, object detection, and situation awareness. LLMs are exceptional at identifying specific objects, individuals, vehicles, or activities in video streams or images, providing detailed insights crucial for military and civilian surveillance operations. It also enables UAVs to operate more autonomously by interpreting and reacting to their surroundings without constant human oversight, significantly benefiting in complex or hostile environments with critical response times.

Moreover, UAVs equipped with LLMs can make real-time decisions about flight paths, areas to focus on, and when to capture critical footage based on the mission's objectives and evolving ground realities. NLP allows UAVs to understand and process human language, enabling them to receive and interpret more complex commands and queries. In addition, LLMs can predict potential security threats or points of interest by analyzing patterns and historical data. This predictive capability allows for proactive surveillance measures, where UAVs can monitor suspected areas more closely or alert human operators about unusual activities or anomalies detected based on learned patterns. It can also enhance real-time decision support by processing and analyzing data on the fly by summarizing vast amounts of collected data into actionable intelligence. It enables quick and informed decisions crucial in surveillance and reconnaissance missions where conditions can change rapidly \cite{kuwertz2018applying}.

\begin{figure*}
  \centering
  \includegraphics[width=18cm]{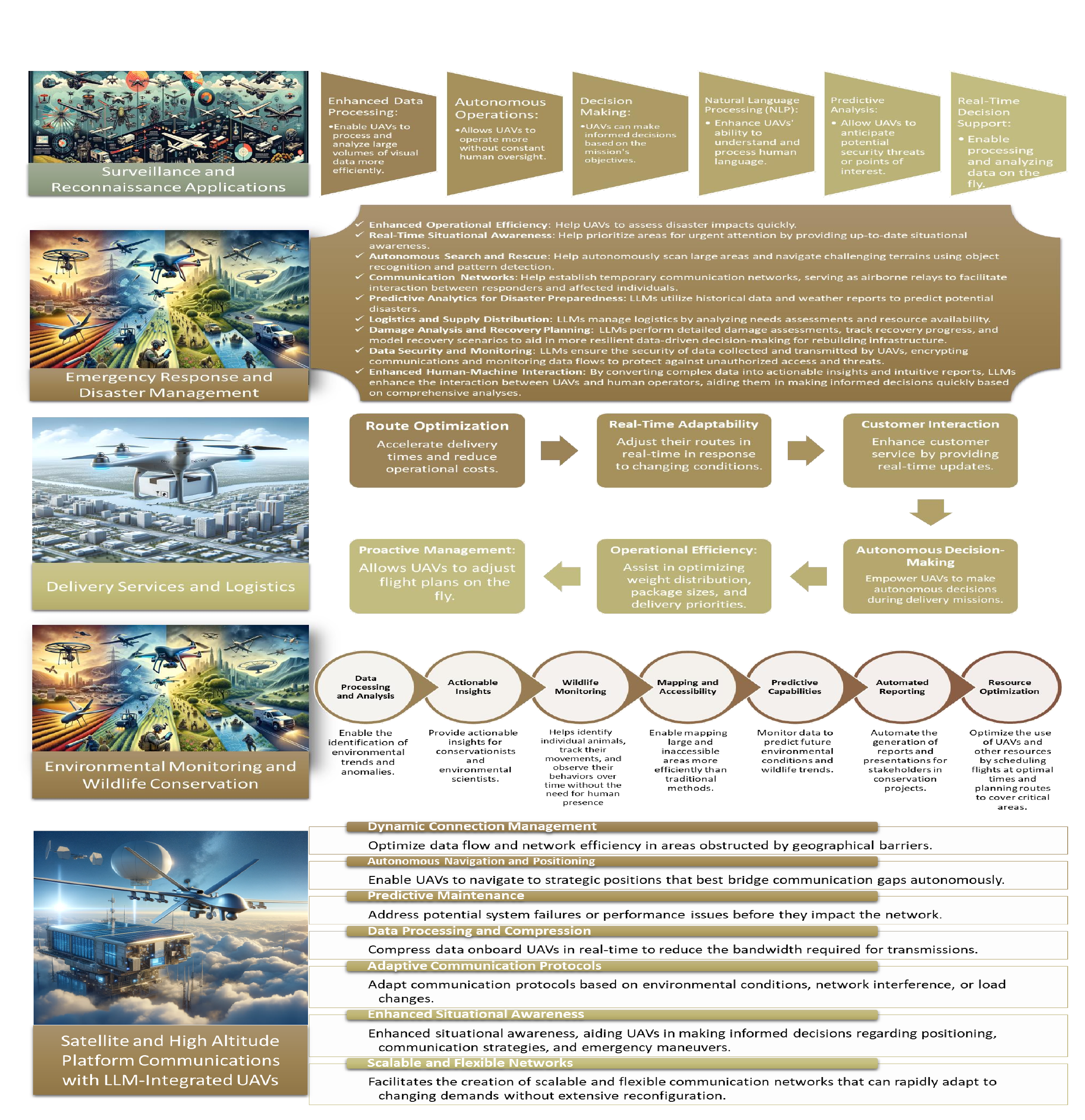}
  \caption{Applications of LLM-integrated UAVs.}
  \label{fig:app}
\end{figure*}

\subsection{Emergency Response and Disaster Management}
LLMs incorporation with UAVs for emergency response and disaster management can significantly enhance the efficiency, accuracy, and effectiveness of emergency operations. LLMs can quickly analyze images and sensor data collected by UAVs to assess the extent of damage immediately following a disaster, including identifying blocked roads, damaged buildings, and flood areas \cite{maharani2020sentiment}. Due to real-time situational awareness, LLM equipped UAVs help emergency responders prioritize areas that require urgent attention and plan the most effective response \cite{goecks2023disasterresponsegpt}.

While in search and rescue missions, time is critical, so UAVs with LLM capabilities can autonomously scan large areas, using object recognition and pattern detection to locate survivors. They can navigate challenging terrains without direct human guidance, speeding up search operations and improving the chances of rescuing distressed people. In addition, as disasters disrupt the communication networks, UAVs integrated with LLMs can establish temporary communication networks, acting as airborne relays to facilitate communication between responders and affected people. LLMs optimize the placement and routing of UAVs to ensure maximum coverage and network efficiency.

Furthermore, they can also improve disaster preparedness due to their predictive analytics abilities by analyzing historical data and current weather reports to predict potential disasters before they occur. This predictive capability allows authorities to deploy UAVs proactively, monitoring at-risk areas and initiating preemptive evacuations or other mitigation measures. Moreover, LLMs can manage logistical aspects by analyzing needs assessments and resource availabilities. They ensure that supplies such as food, water, and medical equipment are optimally distributed and delivered using UAVs, especially to areas that are hard to reach by traditional means due to the disaster \cite{lee2023ai}.

Moreover, they can also play a crucial role in damage analysis and recovery planning by performing detailed assessments of the damage, tracking recovery progress, and analyzing data over time to guide rebuilding efforts. LLMs can model different recovery scenarios, helping planners make data-driven decisions to rebuild infrastructure more resiliently. Given the sensitivity of data involved in emergency response, LLMs ensure that all information collected and transmitted by UAVs is securely encrypted and protected from unauthorized access. They monitor data flows for anomalies that indicate threats, safeguarding critical information during chaotic situations. LLMs enhance interactions between UAVs and human operators by translating complex data into actionable insights and intuitive reports. This allows emergency responders to make informed decisions quickly and effectively based on comprehensive analyzes provided by UAVs in understandable formats.

\subsection{Delivery Services and Logistics}
LLMs-integrated UAVs can transform delivery services and logistics by optimizing routes, enhancing customer interactions, and improving operational efficiency \cite{luo2024language}. LLMs can process complex datasets, including traffic patterns, weather conditions, and geographical data, to dynamically optimize delivery routes. This ensures faster delivery times and helps reduce operational costs. LLMs can adapt these routes in real-time to account for changing conditions, ensuring deliveries are made as efficiently as possible \cite{she2021efficiency}.

UAVs can interact with customers using LMMs to update the status of delivery, answer queries, and even handle complaints or special instructions in real time. Enhanced interaction increases customer satisfaction and streamlines the delivery process, reducing the need for human intervention in customer service. LLMs enable UAVs to make autonomous decisions during delivery missions. For example, when faced with unexpected obstacles or emergencies, UAVs can decide the best course of action, be it rerouting, waiting for clearance, or returning to the base. This level of autonomy ensures reliability and consistency in delivery services, even in unpredictable circumstances. In addition, their proactive approach prevents downtime, extends the lifespan of the UAV fleet, and ensures that technical issues do not disrupt delivery schedules.

Furthermore, LLMs can assist in weight distributions, package sizes, and delivery priorities to ensure that each UAV is loaded efficiently, maximizing delivery capacity and minimizing the number of trips required. LLMs continuously analyze traffic and weather data to adjust UAV flight plans in real-time to maintain delivery schedules, especially in adverse weather conditions or congested airspaces, ensuring that deliveries are made safely and on time.

\subsection{Environmental Monitoring and Wildlife Conservation}
LLMs can process and analyze vast amounts of environmental data collected by UAVs, such as images, temperature readings, and pollution levels. The collected data enable identifying environmental trends and anomalies, such as changes in vegetation cover, water quality, or the presence of pollutants \cite{asadzadeh2022uav,nova2023ai,mashala2023systematic,adu2017water}. LLM can quickly analyze this data and provide actionable insights for conservationists and environmental scientists. LLMs can also help track and study wildlife by analyzing video and audio recordings captured by UAVs to identify individual animals, track their movements, and observe their behaviors over time without human presence, which can reduce stress and behavioral changes in animals caused by human interaction \cite{stephenson2019integrating,chanev2023application}.

Moreover, UAVs integrated with LLMs can map large and inaccessible areas more efficiently than traditional methods. LLMs can analyze the collected geographical data to create detailed maps of habitats, including changes over time. This information is vital for managing natural reserves, planning reforestation projects, or assessing the impacts of human activities on natural habitats. LLMs can also use historical and ongoing monitoring data to predict future environmental conditions and wildlife trends. These predictions can inform conservation efforts, such as predicting the best times and places to implement species protection measures or anticipating ecological changes impacting biodiversity.

LLMs can automate generating reports and presentations for stakeholders involved in environmental conservation projects. By synthesizing complex data into comprehensible formats, LLMs facilitate more transparent communication of findings and recommendations, making it easier for decision-makers to understand the issues and take action. Moreover, LLMs can optimize UAVs and other resources in conservation projects where resources are often limited to ensure maximum coverage and data collection efficiency by scheduling UAV flights at optimal times, planning routes to cover critical areas, and ensuring data is collected cost-effectively.

\subsection{Satellite and High Altitude Platform Communications with LLM-Integrated UAVs}
Integrating LLMs with UAVs for enhancing satellite and High Altitude Platform (HAP) communications involves leveraging advanced analytics and cognitive capabilities to improve data relay, processing, and autonomous decision-making \cite{rong2024leveraging}. Since UAVs serve as mobile nodes or relay points in satellite and HAP communications networks, particularly in areas where direct communication is hindered by geographical barriers or where additional bandwidth is needed temporarily. LLMs can manage these connections dynamically, optimizing data flow between ground stations, satellites, HAPs, and end users. They can make real-time decisions on routing data through UAVs to improve network resilience and reduce latency. LLMs enable UAVs to autonomously navigate to positions where they can most effectively bridge communication gaps between satellites, HAPs, and terrestrial networks. This is particularly useful in disaster areas or during large public events requiring temporary communication boosts.  UAVs, equipped with LLM capabilities, can analyze environmental data, satellite paths, and network demand to determine optimal positions without human intervention \cite{fourati2021artificial}.

LLMs can predict potential system failures or sub-optimal performance before they become critical issues by analyzing telemetry and operational data from UAVs used in satellite and HAP communications. This predictive maintenance capability ensures higher uptime and reliability of the UAVs serving in these critical roles. LLMs can process and compress data onboard UAVs in real-time before relaying it to satellites or HAPs. This reduces the bandwidth needed for data transmission and speeds up communication. The LLMs can employ advanced algorithms to determine the most efficient ways to encode and transmit data based on current network conditions and data priorities.

LLM-integrated UAVs can adapt their communication protocols to maintain effective links with satellites and HAPs in response to changing environmental conditions, interference, or shifting network loads. LLMs can learn from past communications, predict optimal communication windows, and adjust frequencies or modulation schemes to enhance the quality of the connection. In addition, for UAVs operating in complex environments, LLMs provide enhanced situational awareness by processing data from multiple sources, including satellite and HAP sensors. This helps in making informed decisions about UAV positioning, communication strategies, and even emergency maneuvers to avoid conflicts or hazards. Integrating LLMs allows for scalable and flexible communication networks that can adapt to varying demands without extensive reconfiguration. UAVs can be deployed rapidly to scale up network capabilities in response to increased communication needs or to cover temporary satellite or HAP coverage gaps.

\section{Challenges and Considerations in Implementing LLMs-integarted UAVs} \label{challenges}
Implementing LLMs for UAV communication is a novel direction with a series of challenges and considerations that must be addressed to ensure the effectiveness and safety of UAV applications. This section highlights a few fundamental challenges that must be considered to effectively incorporate LLMs in the UAV domain.

\subsection{Computational Resources and Power Consumption}
LLMs require significant computational power and energy to run effectively \cite{10473893,yang2023harnessing}. However, UAVs have limited onboard computing capabilities and power supplies, constrained by the need for lightweight design to ensure longer flight duration and operational efficiency. The power consumption needed to process large models can quickly drain the UAV's batteries, reducing the time available for mission-critical tasks \cite{javaid2023communication,mozaffari2019tutorial}. In addition, adding extra resources can significantly affect UAV efficiency, thus complicating the integration of LLMs. To address these issues, it is crucial to simplify LLMs by pruning unnecessary parameters and using quantization techniques to reduce model size and enhance processing speed with less power consumption \cite{ma2023llm}. Edge computing can further alleviate the need for high-bandwidth connectivity by processing data locally \cite{xu2024cached}.

Furthermore, advanced AI hardware such as Graphics Processing Units (GPUs) \cite{kwon2023efficient}, Field-Programmable Gate Arrays (FPGAs) \cite{chen2024comprehensive}, and model distillation techniques \cite{huang2024leveraging} can help optimize computational demands. Implementing adaptive systems that adjust resource usage based on current needs can also help manage power efficiently, ensuring operational effectiveness without compromising on performance.

\subsection{Communication Latency} The latency challenge in communication is particularly critical when using LLMs for UAV operations involving real-time data processing and decision-making. For example, navigation, surveillance, and tactical responses demand minimal data processing and decision-making delays. However, when LLMs require significant computational resources, a standard solution is to offload this processing to cloud-based servers. While this approach leverages powerful computing capabilities, it inherently introduces latency due to the communication delays from the UAV to the cloud server and back. This delay can be detrimental when crucial immediate responses compromise mission effectiveness and safety \cite{rong2024leveraging}.

To mitigate these issues, UAVs can enhance onboard processing capabilities by integrating advanced computational resources like microprocessors, GPUs, or custom Application-Specific Integrated Circuits (ASICs) to handle complex algorithms more efficiently. Balancing computational power with latency needs is crucial and can be optimized by adopting a hybrid processing approach. This involves handling urgent, real-time processing tasks directly onboard UAV, while delegating more complex, less time-sensitive tasks to the cloud. Such a strategy helps balance the computational load and tailor response times to the urgency and complexity of specific tasks. Additionally, establishing robust near-field communication networks and utilizing edge computing solutions can further reduce latency. By situating processing power closer to the UAV, either through local servers or other nearby UAVs equipped with edge servers, the communication distance and time are significantly decreased, enhancing the overall responsiveness of UAV operations \cite{hassan2023satellite,chen2024survey}.

\subsection{Model Robustness and Reliability} The challenge of model robustness and reliability is critical in deploying UAV communications, as the decisions based on model outputs lead to significant consequences \cite{schwartz2023enhancing}. For example, the models may produce unpredictable or incorrect outputs in novel or edge-case scenarios due to their reliance on patterns learned from training data, which may not adequately cover all possible real-world situations \cite{telli2023comprehensive}. The risk is exceptionally high in dynamic environments where decisions must be made quickly and accurately, as is often the case with UAV operations \cite{de2023semantic,li2024behavior}. Continuously updating and retraining the model with new data can help it learn from recent experiences and adapt to changes or new scenarios it might encounter. This adaptation involves incorporating data from novel situations that UAVs have encountered, expanding the model's understanding and range of responses. For instance, establishing a system where data from UAV missions is regularly fed back into the model's training routine refines and updates its algorithms.

Furthermore, simulation-based testing and validation are essential when relying on LLMs for critical operations. Testing these models under various simulated conditions is crucial to identify potential failures or weaknesses in their responses during complex scenarios such as adverse weather conditions, communication interruptions, or unusual mission parameters. Robust fail-safe mechanisms can also be implemented to prevent harmful actions due to incorrect model outputs by establishing thresholds or conditions requiring human intervention if the model's output is uncertain or falls outside expected parameters. Implementing redundant systems can also double-check critical decisions before execution, and enhanced error handling can address unexpected outputs from LLMs without disrupting the UAV's operations \cite{mishra2023autonomous}.

\subsection{Integration with Existing Systems} Advanced LLMs need to interact seamlessly with the UAV's existing hardware and software modules (such as flight control, navigation systems, communication protocols, and data processing units) each with its unique specifications and operational requirements to enhance decision-making and communication within UAV operations. Integrating LLMs into these diverse frameworks is complex and time-consuming, potentially leading to extensive development and testing periods to ensure full compatibility and functionality. Therefore, adopting a modular approach to system design can significantly ease the integration of LLMs by allowing for the integration, removal, or updating of individual LLM components without disrupting the system. Modular designs offer flexibility and scalability, accommodating the specific needs of different missions or operational adjustments \cite{ullah2024role}.

Furthermore, ensuring the interoperability of new LLM components with existing systems is crucial \cite{wang2024generative}. Interoperability allows different systems and software applications to communicate and work together effectively despite being developed independently. The incremental integration of  LLMs into UAV systems through phased testing and deployment can also reduce the complexity of integration. It also allows for identifying and resolving specific issues without the risk of widespread system failures. In addition, developing a systematic approach for regular updates and maintenance to ensure that the integrated LLMs remain effective and that the overall system adapts to new technological advancements or changes in operational requirements is necessary \cite{telli2023comprehensive}.

\subsection{Data Security and Privacy} Integrating LLMs in UAV operations raises significant concerns regarding data security and privacy, primarily since these models frequently process sensitive data, which may include personal information collected during surveillance missions. This data type is highly vulnerable and, if compromised, can lead to serious privacy violations and other security issues. Implementing robust data security measures is crucial to mitigate these risks. Thus, strong data encryption is fundamental to ensure that data remains inaccessible to unauthorized users while being transmitted and stored \cite{yao2024survey}.

Furthermore, robust access control mechanisms must be established to restrict data access to only authorized personnel, thus preventing any unauthorized data manipulation or leakage. Compliance with data protection regulations is also essential. These regulations are designed to protect the privacy and integrity of data and require organizations to take stringent measures to safeguard all personal information. By adhering to these guidelines, UAV operators can help secure sensitive data processed by LLMs, minimizing the risk of breaches and maintaining the confidentiality and integrity of the information \cite{wu2024new}.
\section{Future Research Directions} \label{future-directions}
This section considers the challenges and considerations previously discussed to outline future research directions. It highlights research areas that require immediate attention to enhance UAVs' intelligence, efficiency, and adaptability through LLM integration. Such exploration is essential for overcoming current limitations and unlocking the full potential of UAV applications across various sectors.

\subsection{Advancements in LLM Algorithms for UAVs}
The future work directions and opportunities for LLM technology advancements for UAV communications are rich and diverse, driven by UAV operations' increasing complexity and demands. Novel schemes should focus on developing LLM algorithms that enable UAVs to dynamically adjust communication protocols and strategies based on real-time data about weather, terrain, and electromagnetic interference. This adaptive capability can significantly improve the effectiveness of UAVs in disaster response and military operations, where conditions can change rapidly and unpredictably \cite{xi2023rise,wu2023autogen}.

Future work should integrate LLMs to enhance UAV swarm intelligence, allowing for sophisticated group behaviors that mimic biological systems. In addition, future research needs to focus on algorithms that enable individual UAVs to make decisions based on the collective input of the swarm, optimizing flight paths and task allocation for efficiency and reduced energy consumption \cite{telli2023comprehensive}. This technology holds promise for applications ranging from large-scale agricultural monitoring to search and rescue missions, where coordinated multi-UAV operations are critical.

In addition, improvement in error correction and signal processing is essential to maintain communication integrity in challenging environments. Future research needs to explore deep learning models to predict and compensate for signal degradation and develop new modulation and coding forms more resistant to interference. This technology can be particularly beneficial in crowded urban areas or during severe weather conditions, where signal loss can critically impair UAV operations \cite{chacko2023paradigm,zhong2024safer}.

Future efforts should also expand the range of applications of LLM-enhanced UAV communications into new fields, such as humanitarian aid, environmental monitoring, and logistics. Future research needs to explore how UAVs equipped with advanced LLM and communication technologies can be deployed in emergency zones to provide real-time updates and aid distribution, monitor wildlife or environmental changes with minimal human involvement, and streamline supply chains with autonomous delivery services.

\subsection{Integration of LLMs with Emerging Technologies}
Integrating LLMs with emerging technologies offers promising advancements for UAV communication systems. For example, incorporating Reconfigurable Intelligent Surfaces (RIS) can optimize signal processing algorithms and significantly enhance the efficiency and reliability of UAV communications by dynamically configuring RIS based on real-time environmental and traffic data \cite{elmossallamy2020reconfigurable}. It can also improve various environments, from smart cities to enhanced healthcare opportunities, by optimizing remote patient monitoring and telemedicine through optimized data transmission. In addition, LLMs can boost the performance of Augmented Reality (AR) and Virtual Reality (VR) applications by supporting high-bandwidth and low-latency communications crucial for immersive experiences \cite{faisal2022machine,yuan2021reconfigurable}.

Furthermore, integrating LLMs with 5G/6G technology can significantly enhance UAV communication capabilities due to the higher bandwidth and lower latency offered by these networks \cite{de2023llm}. It enables UAVs to stream high-definition video for surveillance or inspection tasks, receive updates in real-time for dynamic mission adjustment, and participate in swarm operations with improved coordination. Connecting LLM-equipped UAVs with IoT devices will lead to more interactive and responsive UAV operations within smart cities and industrial environments. UAVs can act as mobile nodes in an IoT network, collecting and processing data from various sources and making decisions on the fly. This integration can be particularly useful in disaster response scenarios, where UAVs can assess damages, detect anomalies, and communicate with other IoT devices to manage emergency services efficiently \cite{ali2021urllc,hong2023llm}.

In addition, integrating LLMs with edge computing platforms can decentralize data processing, reducing the latency involved in cloud computing scenarios and allowing UAVs to perform real-time data analysis at the network's edge. This capability enables UAVs to make quicker decisions during critical missions, such as tracking moving targets or navigating complex terrains without waiting for data to be processed remotely \cite{lin2023pushing}. Similarly, enhancing LLMs with specialized neural networks that can process visual and sensory data can improve UAVs' ability to understand and interact with their environment. It allows UAVs to perform more complex recognition tasks, such as identifying specific individuals in search and rescue operations, detecting structural issues in infrastructure inspection, or monitoring agricultural lands for pest and disease patterns.

Moreover, quantum computing integration can also exponentially increase the processing capabilities of LLMs, enabling them to handle vast datasets more efficiently. Quantum-enhanced LLMs can optimize flight paths and communication protocols far beyond the current capabilities, reducing operational costs and increasing the efficiency of data-heavy tasks \cite{liang2023unleashing}.

\subsection{Computational Efficiency and Power Management Optimization for LLM Integrated-UAV Systems}
To effectively implement LLM-integrated UAV operations, novel schemes should focus on reducing the computational complexity of LLM by removing parameters that do not contribute significantly to the improved performance of UAV communication. Future work can implement a pruning scheme to reduce the model size and computational load, making it more feasible for devices with limited resources. Future schemes should also emphasize on employing quantization techniques that can lower the precision of the model's parameters (e.g., from floating point to integers) to significantly decrease the model size and speed up inference times while consuming less power. In addition, UAVs can also benefit from edge computing services, which allow local data processing without the need to transmit data back to a central server. This reduces the necessity for continuous high-bandwidth connectivity and helps execute complex models by distributing the computational load between the UAV and the edge device.

Furthermore, future hardware design should be explicitly tailored for AI tasks. Employing GPUs. FPGAs or ASICs, optimized for AI interference can significantly enhance power and computational efficiency, delivering superior performance per watt compared to general-purpose processors. Model distillation is another effective strategy that can focused on in the future, and it involves training a smaller 'student' model to replicate the performance of a larger 'teacher' model. The distilled model maintains high accuracy but requires only a fraction of the computational resources, making it suitable for deployment on devices with limited capabilities. Implementing systems that dynamically adjust computing resources based on current needs and available power can optimize power usage. For example, the UAV could deploy a simplified version of the model when battery levels are low or detailed processing is unnecessary \cite{zhu2015using}.
Thus, focusing on these strategies in the future can significantly enhance the feasibility of integrating sophisticated LLMs into UAV systems. These approaches help balance the trade-offs between model performance and the practical limitations of UAV platforms, ensuring that the benefits of advanced NLP capabilities can be harnessed without compromising the operational effectiveness of the UAVs.

\subsection{Latency Reduction Techniques}
In order to address the latency issues, UAVs can enhance their onboard processing power by utilizing advanced computational resources such as microprocessors, GPUs, or custom ASICs, which can efficiently execute complex machine-learning algorithms. Future schemes should consider the trade-off between computational power and latency based on the specific requirements of each UAV mission to address these challenges effectively. A hybrid approach can be particularly effective, where UAVs perform critical real-time processing onboard while more complex but less time-sensitive tasks are offloaded to the cloud. Accordingly, it can balance the computational load and optimize response times according to the urgency and complexity of the tasks. For instance, integrating smart routing algorithms can dynamically determine the best location for processing data, considering current network conditions, task complexity, and the urgency of processing requirements.

Furthermore, future schemes should explore robust near-field communication networks and edge server deployment possibilities to perform computation-intensive tasks at the edge of the network with faster processing speed and lower delay.

\subsection{Standardization Efforts and Protocol Enhancements}
Integrating advanced LLMs into UAV operations presents a complex challenge, as these models must seamlessly interact with a variety of existing UAV hardware and software systems \cite{agapiou2023interacting}. UAV's components, including flight control modules, navigation systems, communication protocols, and data processing units have their own unique specifications and operational demands. This diversity can lead to prolonged development and testing periods needed to ensure full compatibility and functionality.

Future work should focus on adopting a modular system design so individual components can easily be added, removed, or updated without disrupting the overall system integrity \cite{xia2023towards}. Moreover, future efforts must ensure that different systems and software applications can communicate and operate effectively together, even if developed independently. Thus, they can adopt standardized data formats and communication protocols widely accepted within the UAV industry. This helps LLMs understand and adhere to established standards, thereby smoothing the integration process.

In addition, future work should focus on gradually integrating LLMs into UAV systems using a phased approach to ensure compatibility and performance and establishing a systematic framework for regular updates, maintenance, and training by dedicated teams to adapt to technological advancements and maintain effective integration.

\subsection{Reliability Enhancement Strategies}
To improve the reliability of LLM-integrated UAV communication systems, future work should focus on implementing advanced error correction techniques employing robust algorithms to ensure that communication remains reliable even under adverse conditions. The initial testing of LLM based UAV-system should also consider redundancy in communication channels by using multiple communication channels and backup systems to safeguard against the failure of any single channel.

Future work should also focus on AI-driven predictive maintenance by integrating AI tools to predict and schedule maintenance to prevent failures before they occur. It can help minimize downtime and extend the life span of communication components \cite{zhong2024safer}. Furthermore, future schemes should adopt dynamic routing and spectrum management techniques to implement AI-driven dynamic routing algorithms, and spectrum management approaches to optimize available frequencies and paths for data transmission. This approach is beneficial for adapting to changing environmental conditions and communication traffic, enhancing overall system resilience. In addition, AI-based training and simulation of LLM-integrated UAV systems must also be carried out extensively to ensure that they can handle various operational contexts and unexpected situations for enhanced reliability.

Future work should also emphasize the establishment of real-time monitoring and decision support systems \cite{de2023socratic}. These systems are vital as they provide ongoing assessments of UAV health and communication status and can suggest or automate corrective actions.

\subsection{Interference Mitigation Schemes}
Interference mitigation schemes are necessary for LLM-based UAV communication in every domain, ranging from commercial delivery services to essential emergency response operations. To meet these demands, future research must develop advanced signal-processing algorithms that can dynamically identify and mitigate interference in real-time \cite{xu2024penetrative}. This involves employing machine learning models, particularly deep learning techniques that predict and counteract interference patterns based on historical data and real-time inputs \cite{formosa2020predicting,shi2021towards}.

In addition, novel schemes should explore beamforming techniques to improve signal clarity and strength. This can be achieved by implementing smart antenna technologies that adaptively focus and steer beams away from interference sources or use multiple antennas to send and receive signals, thus reducing interference effects \cite{srar2010adaptive,bariah2023large}. Enhancing spectrum management strategies is also vital to optimize frequency usage without causing or suffering from interference. This includes developing LLM-driven models that dynamically allocate bandwidth and adjust frequencies based on the UAV’s mission requirements and the spectral environment.

Future efforts should also focus on integrating cognitive radio capabilities that allow UAV communication systems to automatically change their frequency to avoid interference. Exploring the development of LLM algorithms can enable UAVs to sense their operational environment and make intelligent decisions about frequency hopping or modulation adjustments in essential.

Moreover, improving network coordination among UAVs to manage and mitigate interference collectively is essential. It requires future research into decentralized decision-making models where LLMs enable UAVs to share information about interference sources and collaboratively decide on the best communication paths and protocols. Moreover, strengthening UAV communications against adversarial attacks that can cause interference or disrupt communications is critical. Another key area of focus is utilizing LLMs to develop detection and response systems that identify and neutralize sophisticated signal jamming and spoofing techniques.

\subsection{Regulatory Advocacy and Policy Recommendations}
Future directions and research opportunities for regulatory advocacy and policy recommendations regarding LLM-based UAV communication systems are increasingly pertinent as these technologies become more integrated into various sectors \cite{ghasemi2024accelerating}. The primary focus in the future should be developing comprehensive policies addressing safety, privacy, and ethical standards while promoting innovation and integration in UAV operations. This involves collaborating with regulatory bodies to establish clear guidelines that adapt to the rapid advancements in LLM and UAV technologies.

Future work must ensure secure data communication as UAVs handle and transmit substantial amounts of potentially sensitive data  \cite{pathirana2018towards}. Thus, measures must be implemented to protect this data against breaches and unauthorized access, safeguarding both the integrity of the data and the privacy of individuals \cite{9537930}. In addition, future work should keep focusing on setting airspace usage regulations to prevent conflicts and accidents by determining how UAVs integrate with existing air traffic and defining specific zones or altitudes for UAV operations. At the same time, defining accountability measures for AI decisions is essential as UAVs increasingly make autonomous decisions based on AI; it's crucial to establish who is responsible if those decisions lead to undesirable outcomes. Therefore, creating standards for AI behavior, ensuring AI systems are transparent and their actions are traceable, and establishing legal frameworks to address liability and compliance. Moreover, continuous monitoring and revising of these policies are crucial as technology evolves to sustain an environment that supports technological advancements and protects public interests \cite{al2020comprehensive}.

\section{Conclusion} \label{conclusion}
This paper presents the transformative potential of integrating LLMs with UAVs, ushering in a new era of autonomous systems. We comprehensively analyze LLM architectures,  evaluating their suitability for enhancing UAV capabilities. Our key contributions encompass a detailed assessment of LLM architectures for UAV integration and an exploration of cutting-edge LLM-based UAV architectures. This paves the way for developing significantly more sophisticated, intelligent, and responsive UAV operations.
Furthermore, the focus on improved spectral sensing and sharing through LLM integration unlocks novel avenues for data processing advancements, crucial for robust decision-making within UAV systems. We demonstrate the expanded scope of existing UAV applications by integrating LLMs. We highlight how this empowers them with greater autonomy and more effective responses in various applications, ultimately leading to increased reliability and functionality across diverse sectors.
The paper concludes by outlining critical areas demanding future research to harness the benefits of LLM-UAV integration fully. The advancements discussed lay the groundwork for a future where UAVs transcend their traditional roles, evolving into pivotal components of sophisticated, integrated systems that unlock the full potential of AI. This work can serve as a cornerstone in ongoing technological progress, propelling us toward a future where the synergy between LLMs and UAV technology can revolutionize various domains by achieving unprecedented levels of automation and efficiency.

\bibliographystyle{IEEEtran}
\bibliography{draft}

\begin{thebibliography}{100}
\providecommand{\url}[1]{#1}
\csname url@samestyle\endcsname
\providecommand{\newblock}{\relax}
\providecommand{\bibinfo}[2]{#2}
\providecommand{\BIBentrySTDinterwordspacing}{\spaceskip=0pt\relax}
\providecommand{\BIBentryALTinterwordstretchfactor}{4}
\providecommand{\BIBentryALTinterwordspacing}{\spaceskip=\fontdimen2\font plus
\BIBentryALTinterwordstretchfactor\fontdimen3\font minus
  \fontdimen4\font\relax}
\providecommand{\BIBforeignlanguage}[2]{{%
\expandafter\ifx\csname l@#1\endcsname\relax
\typeout{** WARNING: IEEEtran.bst: No hyphenation pattern has been}%
\typeout{** loaded for the language `#1'. Using the pattern for}%
\typeout{** the default language instead.}%
\else
\language=\csname l@#1\endcsname
\fi
#2}}
\providecommand{\BIBdecl}{\relax}
\BIBdecl

\bibitem{li2021networked}
X.~Li and A.~V. Savkin, ``Networked unmanned aerial vehicles for surveillance
  and monitoring: A survey,'' \emph{Future Internet}, vol.~13, no.~7, p. 174,
  2021.

\bibitem{thakur2021artificial}
N.~Thakur, P.~Nagrath, R.~Jain, D.~Saini, N.~Sharma, and D.~J. Hemanth,
  ``Artificial intelligence techniques in smart cities surveillance using uavs:
  A survey,'' \emph{Machine Intelligence and Data Analytics for Sustainable
  Future Smart Cities}, pp. 329--353, 2021.

\bibitem{ren2019review}
H.~Ren, Y.~Zhao, W.~Xiao, and Z.~Hu, ``A review of uav monitoring in mining
  areas: Current status and future perspectives,'' \emph{International Journal
  of Coal Science and Technology}, vol.~6, pp. 320--333, 2019.

\bibitem{popescu2019survey}
D.~Popescu, F.~Stoican, G.~Stamatescu, O.~Chenaru, and L.~Ichim, ``A survey of
  collaborative uav--wsn systems for efficient monitoring,'' \emph{Sensors},
  vol.~19, no.~21, p. 4690, 2019.

\bibitem{Khalil2023677}
R.~A. Khalil, N.~Saeed, and M.~Almutiry, ``U{AVs-}assisted passive source
  localization using robust {TDOA} ranging for search and rescue,'' \emph{ICT
  Express}, vol.~9, no.~4, pp. 677--682, 2023.

\bibitem{ullah2019uav}
S.~Ullah, K.-I. Kim, K.~H. Kim, M.~Imran, P.~Khan, E.~Tovar, and F.~Ali,
  ``Uav-enabled healthcare architecture: Issues and challenges,'' \emph{Future
  Generation Computer Systems}, vol.~97, pp. 425--432, 2019.

\bibitem{akhtar2023uavs}
M.~W. Akhtar and N.~Saeed, ``U{AVs-E}nabled maritime communications:
  {UAVs-E}nabled maritime communications: Opportunities and challenges,''
  \emph{IEEE Systems, Man, and Cybernetics Magazine}, vol.~9, no.~3, pp. 2--8,
  2023.

\bibitem{mozaffari2019tutorial}
M.~Mozaffari, W.~Saad, M.~Bennis, Y.-H. Nam, and M.~Debbah, ``A tutorial on
  uavs for wireless networks: Applications, challenges, and open problems,''
  \emph{IEEE communications surveys and tutorials}, vol.~21, no.~3, pp.
  2334--2360, 2019.

\bibitem{boroujeni2024comprehensive}
S.~P.~H. Boroujeni, A.~Razi, S.~Khoshdel, F.~Afghah, J.~L. Coen, L.~O’Neill,
  P.~Fule, A.~Watts, N.-M.~T. Kokolakis, and K.~G. Vamvoudakis, ``A
  comprehensive survey of research towards ai-enabled unmanned aerial systems
  in pre-, active-, and post-wildfire management,'' \emph{Information Fusion},
  p. 102369, 2024.

\bibitem{koubaa2023aero}
A.~Koubaa, A.~Ammar, M.~Abdelkader, Y.~Alhabashi, and L.~Ghouti, ``Aero:
  Ai-enabled remote sensing observation with onboard edge computing in uavs,''
  \emph{Remote Sensing}, vol.~15, no.~7, p. 1873, 2023.

\bibitem{cheng2023ai}
N.~Cheng, S.~Wu, X.~Wang, Z.~Yin, C.~Li, W.~Chen, and F.~Chen, ``Ai for
  uav-assisted iot applications: A comprehensive review,'' \emph{IEEE Internet
  of Things Journal}, 2023.

\bibitem{qazi2022iot}
S.~Qazi, B.~A. Khawaja, and Q.~U. Farooq, ``Iot-equipped and ai-enabled next
  generation smart agriculture: A critical review, current challenges and
  future trends,'' \emph{Ieee Access}, vol.~10, pp. 21\,219--21\,235, 2022.

\bibitem{vincent2019sensors}
D.~R. Vincent, N.~Deepa, D.~Elavarasan, K.~Srinivasan, S.~H. Chauhdary, and
  C.~Iwendi, ``Sensors driven ai-based agriculture recommendation model for
  assessing land suitability,'' \emph{Sensors}, vol.~19, no.~17, p. 3667, 2019.

\bibitem{iyer2021ai}
L.~S. Iyer, ``Ai enabled applications towards intelligent transportation,''
  \emph{Transportation Engineering}, vol.~5, p. 100083, 2021.

\bibitem{al2020comprehensive}
F.~Al-Turjman and H.~Zahmatkesh, ``A comprehensive review on the use of ai in
  uav communications: Enabling technologies, applications, and challenges,''
  \emph{Unmanned Aerial Vehicles in Smart Cities}, pp. 1--26, 2020.

\bibitem{Khalil2024}
R.~A. Khalil, Z.~Safelnasr, N.~Yemane, M.~Kedir, A.~Shafiqurrahman, and
  N.~Saeed, ``Advanced learning technologies for intelligent transportation
  systems: {P}rospects and challenges,'' \emph{IEEE Open Journal of Vehicular
  Technology}, vol.~5, pp. 397--427, 2024.

\bibitem{mohammadi2018enabling}
M.~Mohammadi and A.~Al-Fuqaha, ``Enabling cognitive smart cities using big data
  and machine learning: Approaches and challenges,'' \emph{IEEE Communications
  Magazine}, vol.~56, no.~2, pp. 94--101, 2018.

\bibitem{rasley2020deepspeed}
J.~Rasley, S.~Rajbhandari, O.~Ruwase, and Y.~He, ``Deepspeed: System
  optimizations enable training deep learning models with over 100 billion
  parameters,'' in \emph{Proceedings of the 26th ACM SIGKDD International
  Conference on Knowledge Discovery and Data Mining}, 2020, pp. 3505--3506.

\bibitem{ye2023comprehensive}
J.~Ye, X.~Chen, N.~Xu, C.~Zu, Z.~Shao, S.~Liu, Y.~Cui, Z.~Zhou, C.~Gong,
  Y.~Shen \emph{et~al.}, ``A comprehensive capability analysis of gpt-3 and
  gpt-3.5 series models,'' \emph{arXiv preprint arXiv:2303.10420}, 2023.

\bibitem{devlin2018bert}
J.~Devlin, M.-W. Chang, K.~Lee, and K.~Toutanova, ``Bert: Pre-training of deep
  bidirectional transformers for language understanding,'' \emph{arXiv preprint
  arXiv:1810.04805}, 2018.

\bibitem{mastropaolo2021studying}
A.~Mastropaolo, S.~Scalabrino, N.~Cooper, D.~N. Palacio, D.~Poshyvanyk,
  R.~Oliveto, and G.~Bavota, ``Studying the usage of text-to-text transfer
  transformer to support code-related tasks,'' in \emph{2021 IEEE/ACM 43rd
  International Conference on Software Engineering (ICSE)}.\hskip 1em plus
  0.5em minus 0.4em\relax IEEE, 2021, pp. 336--347.

\bibitem{liu2024generative}
G.~Liu, N.~Van~Huynh, H.~Du, D.~T. Hoang, D.~Niyato, K.~Zhu, J.~Kang, Z.~Xiong,
  A.~Jamalipour, and D.~I. Kim, ``Generative ai for unmanned vehicle swarms:
  Challenges, applications and opportunities,'' \emph{arXiv preprint
  arXiv:2402.18062}, 2024.

\bibitem{agapiou2023interacting}
A.~Agapiou and V.~Lysandrou, ``Interacting with the artificial intelligence
  (ai) language model chatgpt: a synopsis of earth observation and remote
  sensing in archaeology,'' \emph{Heritage}, vol.~6, no.~5, pp. 4072--4085,
  2023.

\bibitem{kurunathan2023machine}
H.~Kurunathan, H.~Huang, K.~Li, W.~Ni, and E.~Hossain, ``Machine learning-aided
  operations and communications of unmanned aerial vehicles: A contemporary
  survey,'' \emph{IEEE Communications Surveys and Tutorials}, 2023.

\bibitem{rong2024leveraging}
B.~Rong and H.~Rutagemwa, ``Leveraging large language models for intelligent
  control of 6g integrated tn-ntn with iot service,'' \emph{IEEE Network},
  2024.

\bibitem{ullah2024role}
A.~Ullah, G.~Qi, S.~Hussain, I.~Ullah, and Z.~Ali, ``The role of llms in
  sustainable smart cities: Applications, challenges, and future directions,''
  \emph{arXiv preprint arXiv:2402.14596}, 2024.

\bibitem{liu2023aerialvln}
S.~Liu, H.~Zhang, Y.~Qi, P.~Wang, Y.~Zhang, and Q.~Wu, ``Aerialvln:
  Vision-and-language navigation for uavs,'' in \emph{Proceedings of the
  IEEE/CVF International Conference on Computer Vision}, 2023, pp.
  15\,384--15\,394.

\bibitem{piggott2023net}
B.~Piggott, S.~Patil, G.~Feng, I.~Odat, R.~Mukherjee, B.~Dharmalingam, and
  A.~Liu, ``Net-gpt: A llm-empowered man-in-the-middle chatbot for unmanned
  aerial vehicle,'' in \emph{2023 IEEE/ACM Symposium on Edge Computing
  (SEC)}.\hskip 1em plus 0.5em minus 0.4em\relax IEEE, 2023, pp. 287--293.

\bibitem{de2023socratic}
I.~De~Zarz{\`a}, J.~de~Curt{\`o}, and C.~T. Calafate, ``Socratic video
  understanding on unmanned aerial vehicles,'' \emph{Procedia Computer
  Science}, vol. 225, pp. 144--154, 2023.

\bibitem{sun2024generative}
G.~Sun, W.~Xie, D.~Niyato, H.~Du, J.~Kang, J.~Wu, S.~Sun, and P.~Zhang,
  ``Generative ai for advanced uav networking,'' \emph{arXiv preprint
  arXiv:2404.10556}, 2024.

\bibitem{yun2021attention}
W.~J. Yun, B.~Lim, S.~Jung, Y.-C. Ko, J.~Park, J.~Kim, and M.~Bennis,
  ``Attention-based reinforcement learning for real-time uav semantic
  communication,'' in \emph{2021 17th International Symposium on Wireless
  Communication Systems (ISWCS)}.\hskip 1em plus 0.5em minus 0.4em\relax IEEE,
  2021, pp. 1--6.

\bibitem{zhao2024expel}
A.~Zhao, D.~Huang, Q.~Xu, M.~Lin, Y.-J. Liu, and G.~Huang, ``Expel: Llm agents
  are experiential learners,'' in \emph{Proceedings of the AAAI Conference on
  Artificial Intelligence}, vol.~38, no.~17, 2024, pp. 19\,632--19\,642.

\bibitem{eigner2024determinants}
E.~Eigner and T.~H{\"a}ndler, ``Determinants of llm-assisted decision-making,''
  \emph{arXiv preprint arXiv:2402.17385}, 2024.

\bibitem{jin2023time}
M.~Jin, S.~Wang, L.~Ma, Z.~Chu, J.~Y. Zhang, X.~Shi, P.-Y. Chen, Y.~Liang,
  Y.-F. Li, S.~Pan \emph{et~al.}, ``Time-llm: Time series forecasting by
  reprogramming large language models,'' \emph{arXiv preprint
  arXiv:2310.01728}, 2023.

\bibitem{zhu2024llms}
J.~Zhu, S.~Cai, F.~Deng, and J.~Wu, ``Do llms understand visual anomalies?
  uncovering llm capabilities in zero-shot anomaly detection,'' \emph{arXiv
  preprint arXiv:2404.09654}, 2024.

\bibitem{wang2024survey}
L.~Wang, C.~Ma, X.~Feng, Z.~Zhang, H.~Yang, J.~Zhang, Z.~Chen, J.~Tang,
  X.~Chen, Y.~Lin \emph{et~al.}, ``A survey on large language model based
  autonomous agents,'' \emph{Frontiers of Computer Science}, vol.~18, no.~6,
  pp. 1--26, 2024.

\bibitem{xi2023rise}
Z.~Xi, W.~Chen, X.~Guo, W.~He, Y.~Ding, B.~Hong, M.~Zhang, J.~Wang, S.~Jin,
  E.~Zhou \emph{et~al.}, ``The rise and potential of large language model based
  agents: A survey,'' \emph{arXiv preprint arXiv:2309.07864}, 2023.

\bibitem{wang2023aligning}
Y.~Wang, W.~Zhong, L.~Li, F.~Mi, X.~Zeng, W.~Huang, L.~Shang, X.~Jiang, and
  Q.~Liu, ``Aligning large language models with human: A survey,'' \emph{arXiv
  preprint arXiv:2307.12966}, 2023.

\bibitem{zhu2023survey}
X.~Zhu, J.~Li, Y.~Liu, C.~Ma, and W.~Wang, ``A survey on model compression for
  large language models,'' \emph{arXiv preprint arXiv:2308.07633}, 2023.

\bibitem{gao2024llm}
M.~Gao, X.~Hu, J.~Ruan, X.~Pu, and X.~Wan, ``Llm-based nlg evaluation: Current
  status and challenges,'' \emph{arXiv preprint arXiv:2402.01383}, 2024.

\bibitem{kaddour2023challenges}
J.~Kaddour, J.~Harris, M.~Mozes, H.~Bradley, R.~Raileanu, and R.~McHardy,
  ``Challenges and applications of large language models,'' \emph{arXiv
  preprint arXiv:2307.10169}, 2023.

\bibitem{bariah2024large}
L.~Bariah, Q.~Zhao, H.~Zou, Y.~Tian, F.~Bader, and M.~Debbah, ``Large
  generative ai models for telecom: The next big thing?'' \emph{IEEE
  Communications Magazine}, 2024.

\bibitem{du2023power}
Y.~Du, S.~C. Liew, K.~Chen, and Y.~Shao, ``The power of large language models
  for wireless communication system development: A case study on fpga
  platforms,'' \emph{arXiv preprint arXiv:2307.07319}, 2023.

\bibitem{gao2023retrieval}
Y.~Gao, Y.~Xiong, X.~Gao, K.~Jia, J.~Pan, Y.~Bi, Y.~Dai, J.~Sun, and H.~Wang,
  ``Retrieval-augmented generation for large language models: A survey,''
  \emph{arXiv preprint arXiv:2312.10997}, 2023.

\bibitem{mialon2023augmented}
G.~Mialon, R.~Dess{\`\i}, M.~Lomeli, C.~Nalmpantis, R.~Pasunuru, R.~Raileanu,
  B.~Rozi{\`e}re, T.~Schick, J.~Dwivedi-Yu, A.~Celikyilmaz \emph{et~al.},
  ``Augmented language models: a survey,'' \emph{arXiv preprint
  arXiv:2302.07842}, 2023.

\bibitem{schwartz2023enhancing}
S.~Schwartz, A.~Yaeli, and S.~Shlomov, ``Enhancing trust in llm-based ai
  automation agents: New considerations and future challenges,'' \emph{arXiv
  preprint arXiv:2308.05391}, 2023.

\bibitem{yin2023survey}
S.~Yin, C.~Fu, S.~Zhao, K.~Li, X.~Sun, T.~Xu, and E.~Chen, ``A survey on
  multimodal large language models,'' \emph{arXiv preprint arXiv:2306.13549},
  2023.

\bibitem{zhang2023instruction}
S.~Zhang, L.~Dong, X.~Li, S.~Zhang, X.~Sun, S.~Wang, J.~Li, R.~Hu, T.~Zhang,
  F.~Wu \emph{et~al.}, ``Instruction tuning for large language models: A
  survey,'' \emph{arXiv preprint arXiv:2308.10792}, 2023.

\bibitem{huang2022towards}
J.~Huang and K.~C.-C. Chang, ``Towards reasoning in large language models: A
  survey,'' \emph{arXiv preprint arXiv:2212.10403}, 2022.

\bibitem{liu2024large}
X.~Liu, P.~Xu, J.~Wu, J.~Yuan, Y.~Yang, Y.~Zhou, F.~Liu, T.~Guan, H.~Wang,
  T.~Yu \emph{et~al.}, ``Large language models and causal inference in
  collaboration: A comprehensive survey,'' \emph{arXiv preprint
  arXiv:2403.09606}, 2024.

\bibitem{jozefowicz2016exploring}
R.~Jozefowicz, O.~Vinyals, M.~Schuster, N.~Shazeer, and Y.~Wu, ``Exploring the
  limits of language modeling,'' \emph{arXiv preprint arXiv:1602.02410}, 2016.

\bibitem{zhang2023automl}
S.~Zhang, C.~Gong, L.~Wu, X.~Liu, and M.~Zhou, ``Automl-gpt: Automatic machine
  learning with gpt,'' \emph{arXiv preprint arXiv:2305.02499}, 2023.

\bibitem{peng2024limitations}
B.~Peng, S.~Narayanan, and C.~Papadimitriou, ``On limitations of the
  transformer architecture,'' \emph{arXiv preprint arXiv:2402.08164}, 2024.

\bibitem{naveed2023comprehensive}
H.~Naveed, A.~U. Khan, S.~Qiu, M.~Saqib, S.~Anwar, M.~Usman, N.~Barnes, and
  A.~Mian, ``A comprehensive overview of large language models,'' \emph{arXiv
  preprint arXiv:2307.06435}, 2023.

\bibitem{vaswani2017attention}
A.~Vaswani, N.~Shazeer, N.~Parmar, J.~Uszkoreit, L.~Jones, A.~N. Gomez,
  {\L}.~Kaiser, and I.~Polosukhin, ``Attention is all you need,''
  \emph{Advances in neural information processing systems}, vol.~30, 2017.

\bibitem{alaparthi2020bidirectional}
S.~Alaparthi and M.~Mishra, ``Bidirectional encoder representations from
  transformers (bert): A sentiment analysis odyssey,'' \emph{arXiv preprint
  arXiv:2007.01127}, 2020.

\bibitem{sun2019bert4rec}
F.~Sun, J.~Liu, J.~Wu, C.~Pei, X.~Lin, W.~Ou, and P.~Jiang, ``Bert4rec:
  Sequential recommendation with bidirectional encoder representations from
  transformer,'' in \emph{Proceedings of the 28th ACM international conference
  on information and knowledge management}, 2019, pp. 1441--1450.

\bibitem{sun2019ernie}
Y.~Sun, S.~Wang, Y.~Li, S.~Feng, X.~Chen, H.~Zhang, X.~Tian, D.~Zhu, H.~Tian,
  and H.~Wu, ``Ernie: Enhanced representation through knowledge integration,''
  \emph{arXiv preprint arXiv:1904.09223}, 2019.

\bibitem{rosario2024generative}
A.~T. Ros{\'a}rio, ``Generative ai and generative pre-trained transformer
  applications: Challenges and opportunities,'' \emph{Making Art With
  Generative AI Tools}, pp. 45--71, 2024.

\bibitem{yenduri2024gpt}
G.~Yenduri, M.~Ramalingam, G.~C. Selvi, Y.~Supriya, G.~Srivastava, P.~K.~R.
  Maddikunta, G.~D. Raj, R.~H. Jhaveri, B.~Prabadevi, W.~Wang \emph{et~al.},
  ``Gpt (generative pre-trained transformer)--a comprehensive review on
  enabling technologies, potential applications, emerging challenges, and
  future directions,'' \emph{IEEE Access}, 2024.

\bibitem{rodriguez2022end}
R.~Rodriguez-Torrealba, E.~Garcia-Lopez, and A.~Garcia-Cabot, ``End-to-end
  generation of multiple-choice questions using text-to-text transfer
  transformer models,'' \emph{Expert Systems with Applications}, vol. 208, p.
  118258, 2022.

\bibitem{yang2019xlnet}
Z.~Yang, Z.~Dai, Y.~Yang, J.~Carbonell, R.~R. Salakhutdinov, and Q.~V. Le,
  ``Xlnet: Generalized autoregressive pretraining for language understanding,''
  \emph{Advances in neural information processing systems}, vol.~32, 2019.

\bibitem{sankar2022comparative}
A.~Sankar and R.~Dhanalakshmi, ``Comparative study of transformer models,'' in
  \emph{Australasian Database Conference}.\hskip 1em plus 0.5em minus
  0.4em\relax Springer, 2022, pp. 193--200.

\bibitem{de2023semantic}
J.~De~Curt{\`o}, I.~De~Zarz{\`a}, and C.~T. Calafate, ``Semantic scene
  understanding with large language models on unmanned aerial vehicles,''
  \emph{Drones}, vol.~7, no.~2, p. 114, 2023.

\bibitem{ONEATA2021106943}
\BIBentryALTinterwordspacing
D.~Oneață and H.~Cucu, ``Multimodal speech recognition for unmanned aerial
  vehicles,'' \emph{Computers and Electrical Engineering}, vol.~90, p. 106943,
  2021. [Online]. Available:
  \url{https://www.sciencedirect.com/science/article/pii/S0045790620307904}
\BIBentrySTDinterwordspacing

\bibitem{choutri2022multi}
K.~Choutri, M.~Lagha, S.~Meshoul, M.~Batouche, Y.~Kacel, and N.~Mebarkia, ``A
  multi-lingual speech recognition-based framework to human-drone
  interaction,'' \emph{Electronics}, vol.~11, no.~12, p. 1829, 2022.

\bibitem{tagliabue2023real}
A.~Tagliabue, K.~Kondo, T.~Zhao, M.~Peterson, C.~T. Tewari, and J.~P. How,
  ``Real: Resilience and adaptation using large language models on autonomous
  aerial robots,'' \emph{arXiv preprint arXiv:2311.01403}, 2023.

\bibitem{zhong2024safer}
J.~Zhong, M.~Li, Y.~Chen, Z.~Wei, F.~Yang, and H.~Shen, ``A safer vision-based
  autonomous planning system for quadrotor uavs with dynamic obstacle
  trajectory prediction and its application with llms,'' in \emph{Proceedings
  of the IEEE/CVF Winter Conference on Applications of Computer Vision}, 2024,
  pp. 920--929.

\bibitem{jawhar2017communication}
I.~Jawhar, N.~Mohamed, J.~Al-Jaroodi, D.~P. Agrawal, and S.~Zhang,
  ``Communication and networking of uav-based systems: Classification and
  associated architectures,'' \emph{Journal of Network and Computer
  Applications}, vol.~84, pp. 93--108, 2017.

\bibitem{koroteev2021bert}
M.~Koroteev, ``Bert: a review of applications in natural language processing
  and understanding,'' \emph{arXiv preprint arXiv:2103.11943}, 2021.

\bibitem{ravichandiran2021getting}
S.~Ravichandiran, \emph{Getting Started with Google BERT: Build and train
  state-of-the-art natural language processing models using BERT}.\hskip 1em
  plus 0.5em minus 0.4em\relax Packt Publishing Ltd, 2021.

\bibitem{ehrmann2023named}
M.~Ehrmann, A.~Hamdi, E.~L. Pontes, M.~Romanello, and A.~Doucet, ``Named entity
  recognition and classification in historical documents: A survey,'' \emph{ACM
  Computing Surveys}, vol.~56, no.~2, pp. 1--47, 2023.

\bibitem{hakala2019biomedical}
K.~Hakala and S.~Pyysalo, ``Biomedical named entity recognition with
  multilingual bert,'' in \emph{Proceedings of the 5th workshop on BioNLP open
  shared tasks}, 2019, pp. 56--61.

\bibitem{xu2019bert}
H.~Xu, B.~Liu, L.~Shu, and P.~S. Yu, ``Bert post-training for review reading
  comprehension and aspect-based sentiment analysis,'' \emph{arXiv preprint
  arXiv:1904.02232}, 2019.

\bibitem{alaparthi2021bert}
S.~Alaparthi and M.~Mishra, ``Bert: A sentiment analysis odyssey,''
  \emph{Journal of Marketing Analytics}, vol.~9, no.~2, pp. 118--126, 2021.

\bibitem{liu2019roberta}
Y.~Liu, M.~Ott, N.~Goyal, J.~Du, M.~Joshi, D.~Chen, O.~Levy, M.~Lewis,
  L.~Zettlemoyer, and V.~Stoyanov, ``Roberta: A robustly optimized bert
  pretraining approach,'' \emph{arXiv preprint arXiv:1907.11692}, 2019.

\bibitem{sanh2019distilbert}
V.~Sanh, L.~Debut, J.~Chaumond, and T.~Wolf, ``Distilbert, a distilled version
  of bert: smaller, faster, cheaper and lighter,'' \emph{arXiv preprint
  arXiv:1910.01108}, 2019.

\bibitem{lan2019albert}
Z.~Lan, M.~Chen, S.~Goodman, K.~Gimpel, P.~Sharma, and R.~Soricut, ``Albert: A
  lite bert for self-supervised learning of language representations,''
  \emph{arXiv preprint arXiv:1909.11942}, 2019.

\bibitem{luo2024language}
S.~Luo, Y.~Yao, H.~Zhao, and L.~Song, ``A language model-based fine-grained
  address resolution framework in uav delivery system,'' \emph{IEEE Journal of
  Selected Topics in Signal Processing}, 2024.

\bibitem{silalahi2022named}
S.~Silalahi, T.~Ahmad, and H.~Studiawan, ``Named entity recognition for drone
  forensic using bert and distilbert,'' in \emph{2022 International Conference
  on Data Science and Its Applications (ICoDSA)}.\hskip 1em plus 0.5em minus
  0.4em\relax IEEE, 2022, pp. 53--58.

\bibitem{silalahi2023transformer}
------, ``Transformer-based named entity recognition on drone flight logs to
  support forensic investigation,'' \emph{IEEE Access}, vol.~11, pp.
  3257--3274, 2023.

\bibitem{fan2023research}
Y.~Fan, B.~Mi, Y.~Sun, and L.~Yin, ``Research on the intelligent construction
  of uav knowledge graph based on attentive semantic representation,''
  \emph{Drones}, vol.~7, no.~6, p. 360, 2023.

\bibitem{radford2018improving}
A.~Radford, K.~Narasimhan, T.~Salimans, I.~Sutskever \emph{et~al.}, ``Improving
  language understanding by generative pre-training,'' 2018.

\bibitem{radford2019language}
A.~Radford, J.~Wu, R.~Child, D.~Luan, D.~Amodei, I.~Sutskever \emph{et~al.},
  ``Language models are unsupervised multitask learners,'' \emph{OpenAI blog},
  vol.~1, no.~8, p.~9, 2019.

\bibitem{kalyan2023survey}
K.~S. Kalyan, ``A survey of gpt-3 family large language models including
  chatgpt and gpt-4,'' \emph{Natural Language Processing Journal}, p. 100048,
  2023.

\bibitem{brown2020language}
T.~Brown, B.~Mann, N.~Ryder, M.~Subbiah, J.~D. Kaplan, P.~Dhariwal,
  A.~Neelakantan, P.~Shyam, G.~Sastry, A.~Askell \emph{et~al.}, ``Language
  models are few-shot learners,'' \emph{Advances in neural information
  processing systems}, vol.~33, pp. 1877--1901, 2020.

\bibitem{hendy2023good}
A.~Hendy, M.~Abdelrehim, A.~Sharaf, V.~Raunak, M.~Gabr, H.~Matsushita, Y.~J.
  Kim, M.~Afify, and H.~H. Awadalla, ``How good are gpt models at machine
  translation? a comprehensive evaluation,'' \emph{arXiv preprint
  arXiv:2302.09210}, 2023.

\bibitem{achiam2023gpt}
J.~Achiam, S.~Adler, S.~Agarwal, L.~Ahmad, I.~Akkaya, F.~L. Aleman, D.~Almeida,
  J.~Altenschmidt, S.~Altman, S.~Anadkat \emph{et~al.}, ``Gpt-4 technical
  report,'' \emph{arXiv preprint arXiv:2303.08774}, 2023.

\bibitem{tazir2023words}
M.~L. TAZIR, M.~MANCAS, and T.~DUTOIT, ``From words to flight: Integrating
  openai chatgpt with px4/gazebo for natural language-based drone control,'' in
  \emph{International Workshop on Computer Science and Engineering}, 2023.

\bibitem{wang2017construction}
S.~Wang, J.~Chen, Z.~Zhang, G.~Wang, Y.~Tan, and Y.~Zheng, ``Construction of a
  virtual reality platform for uav deep learning,'' in \emph{2017 Chinese
  Automation Congress (CAC)}.\hskip 1em plus 0.5em minus 0.4em\relax IEEE,
  2017, pp. 3912--3916.

\bibitem{biswas2023prospective}
S.~Biswas, ``Prospective role of chat gpt in the military: According to
  chatgpt,'' \emph{Qeios}, 2023.

\bibitem{choudhury2024natural}
N.~R. Choudhury, Y.~Wen, and K.~Chen, ``Natural language navigation for robotic
  systems: Integrating gpt and dense captioning models with object detection in
  autonomous inspections,'' in \emph{Construction Research Congress 2024}, pp.
  972--980.

\bibitem{wang2023survey}
Y.~Wang, Y.~Pan, M.~Yan, Z.~Su, and T.~H. Luan, ``A survey on chatgpt:
  Ai-generated contents, challenges, and solutions,'' \emph{IEEE Open Journal
  of the Computer Society}, 2023.

\bibitem{raffel2020exploring}
C.~Raffel, N.~Shazeer, A.~Roberts, K.~Lee, S.~Narang, M.~Matena, Y.~Zhou,
  W.~Li, and P.~J. Liu, ``Exploring the limits of transfer learning with a
  unified text-to-text transformer,'' \emph{Journal of machine learning
  research}, vol.~21, no. 140, pp. 1--67, 2020.

\bibitem{ni2021sentence}
J.~Ni, G.~H. Abrego, N.~Constant, J.~Ma, K.~B. Hall, D.~Cer, and Y.~Yang,
  ``Sentence-t5: Scalable sentence encoders from pre-trained text-to-text
  models,'' \emph{arXiv preprint arXiv:2108.08877}, 2021.

\bibitem{48643}
A.~Roberts, C.~Raffel, K.~Lee, M.~Matena, N.~Shazeer, P.~J. Liu, S.~Narang,
  W.~Li, and Y.~Zhou, ``Exploring the limits of transfer learning with a
  unified text-to-text transformer,'' Google, Tech. Rep., 2019.

\bibitem{cortiz2021exploring}
D.~Cortiz, ``Exploring transformers in emotion recognition: a comparison of
  bert, distillbert, roberta, xlnet and electra,'' \emph{arXiv preprint
  arXiv:2104.02041}, 2021.

\bibitem{adoma2020comparative}
A.~F. Adoma, N.-M. Henry, and W.~Chen, ``Comparative analyses of bert, roberta,
  distilbert, and xlnet for text-based emotion recognition,'' in \emph{2020
  17th International Computer Conference on Wavelet Active Media Technology and
  Information Processing (ICCWAMTIP)}.\hskip 1em plus 0.5em minus 0.4em\relax
  IEEE, 2020, pp. 117--121.

\bibitem{rajapaksha2021bert}
P.~Rajapaksha, R.~Farahbakhsh, and N.~Crespi, ``Bert, xlnet or roberta: the
  best transfer learning model to detect clickbaits,'' \emph{IEEE Access},
  vol.~9, pp. 154\,704--154\,716, 2021.

\bibitem{li2020comparing}
H.~Li, J.~Choi, S.~Lee, and J.~H. Ahn, ``Comparing bert and xlnet from the
  perspective of computational characteristics,'' in \emph{2020 International
  Conference on Electronics, Information, and Communication (ICEIC)}.\hskip 1em
  plus 0.5em minus 0.4em\relax IEEE, 2020, pp. 1--4.

\bibitem{topal2021exploring}
M.~O. Topal, A.~Bas, and I.~van Heerden, ``Exploring transformers in natural
  language generation: Gpt, bert, and xlnet,'' \emph{arXiv preprint
  arXiv:2102.08036}, 2021.

\bibitem{oneata2019kite}
D.~Oneata and H.~Cucu, ``Kite: Automatic speech recognition for unmanned aerial
  vehicles,'' \emph{arXiv preprint arXiv:1907.01195}, 2019.

\bibitem{li2024behavior}
L.~Li, R.~Yang, M.~Lv, A.~Wu, and Z.~Zhao, ``From behavior to natural language:
  Generative approach for unmanned aerial vehicle intent recognition,''
  \emph{IEEE Transactions on Artificial Intelligence}, 2024.

\bibitem{yao2024can}
Y.~Yao, S.~Luo, H.~Zhao, G.~Deng, and L.~Song, ``Can llm substitute human
  labeling? a case study of fine-grained chinese address entity recognition
  dataset for uav delivery,'' \emph{arXiv preprint arXiv:2403.06097}, 2024.

\bibitem{sun2020ernie}
Y.~Sun, S.~Wang, Y.~Li, S.~Feng, H.~Tian, H.~Wu, and H.~Wang, ``Ernie 2.0: A
  continual pre-training framework for language understanding,'' in
  \emph{Proceedings of the AAAI conference on artificial intelligence},
  vol.~34, no.~05, 2020, pp. 8968--8975.

\bibitem{sun2021ernie}
Y.~Sun, S.~Wang, S.~Feng, S.~Ding, C.~Pang, J.~Shang, J.~Liu, X.~Chen, Y.~Zhao,
  Y.~Lu \emph{et~al.}, ``Ernie 3.0: Large-scale knowledge enhanced pre-training
  for language understanding and generation,'' \emph{arXiv preprint
  arXiv:2107.02137}, 2021.

\bibitem{xiao2020ernie}
D.~Xiao, H.~Zhang, Y.~Li, Y.~Sun, H.~Tian, H.~Wu, and H.~Wang, ``Ernie-gen: An
  enhanced multi-flow pre-training and fine-tuning framework for natural
  language generation,'' \emph{arXiv preprint arXiv:2001.11314}, 2020.

\bibitem{yu2021ernie}
F.~Yu, J.~Tang, W.~Yin, Y.~Sun, H.~Tian, H.~Wu, and H.~Wang, ``Ernie-vil:
  Knowledge enhanced vision-language representations through scene graphs,'' in
  \emph{Proceedings of the AAAI conference on artificial intelligence},
  vol.~35, no.~4, 2021, pp. 3208--3216.

\bibitem{lewis2019bart}
M.~Lewis, Y.~Liu, N.~Goyal, M.~Ghazvininejad, A.~Mohamed, O.~Levy, V.~Stoyanov,
  and L.~Zettlemoyer, ``Bart: Denoising sequence-to-sequence pre-training for
  natural language generation, translation, and comprehension,'' \emph{arXiv
  preprint arXiv:1910.13461}, 2019.

\bibitem{bhattacharjee2020bert}
K.~Bhattacharjee, M.~Ballesteros, R.~Anubhai, S.~Muresan, J.~Ma, F.~Ladhak, and
  Y.~Al-Onaizan, ``To bert or not to bert: Comparing task-specific and
  task-agnostic semi-supervised approaches for sequence tagging,'' \emph{arXiv
  preprint arXiv:2010.14042}, 2020.

\bibitem{van2011cognitive}
D.~Van~Ravenzwaaij, G.~Dutilh, and E.-J. Wagenmakers, ``Cognitive model
  decomposition of the bart: Assessment and application,'' \emph{Journal of
  Mathematical Psychology}, vol.~55, no.~1, pp. 94--105, 2011.

\bibitem{wei2020performance}
Z.~Wei, J.~Zhu, Z.~Guo, and F.~Ning, ``The performance analysis of spectrum
  sharing between uav enabled wireless mesh networks and ground networks,''
  \emph{IEEE Sensors Journal}, vol.~21, no.~5, pp. 7034--7045, 2020.

\bibitem{jasim2021survey}
M.~A. Jasim, H.~Shakhatreh, N.~Siasi, A.~H. Sawalmeh, A.~Aldalbahi, and
  A.~Al-Fuqaha, ``A survey on spectrum management for unmanned aerial vehicles
  (uavs),'' \emph{IEEE Access}, vol.~10, pp. 11\,443--11\,499, 2021.

\bibitem{wang2018enabling}
L.~Wang, H.~Yang, J.~Long, K.~Wu, and J.~Chen, ``Enabling ultra-dense uav-aided
  network with overlapped spectrum sharing: Potential and approaches,''
  \emph{IEEE Network}, vol.~32, no.~5, pp. 85--91, 2018.

\bibitem{shamsoshoara2019distributed}
A.~Shamsoshoara, M.~Khaledi, F.~Afghah, A.~Razi, and J.~Ashdown, ``Distributed
  cooperative spectrum sharing in uav networks using multi-agent reinforcement
  learning,'' in \emph{2019 16th IEEE Annual Consumer Communications \&
  Networking Conference (CCNC)}.\hskip 1em plus 0.5em minus 0.4em\relax IEEE,
  2019, pp. 1--6.

\bibitem{shang2020spectrum}
B.~Shang, V.~Marojevic, Y.~Yi, A.~S. Abdalla, and L.~Liu, ``Spectrum sharing
  for uav communications: Spatial spectrum sensing and open issues,''
  \emph{IEEE Vehicular Technology Magazine}, vol.~15, no.~2, pp. 104--112,
  2020.

\bibitem{zhang2016spectrum}
C.~Zhang and W.~Zhang, ``Spectrum sharing for drone networks,'' \emph{IEEE
  Journal on Selected Areas in Communications}, vol.~35, no.~1, pp. 136--144,
  2016.

\bibitem{lin2020dynamic}
Y.~Lin, M.~Wang, X.~Zhou, G.~Ding, and S.~Mao, ``Dynamic spectrum interaction
  of uav flight formation communication with priority: A deep reinforcement
  learning approach,'' \emph{IEEE Transactions on Cognitive Communications and
  Networking}, vol.~6, no.~3, pp. 892--903, 2020.

\bibitem{massaro2017next}
M.~Massaro, ``Next generation of radio spectrum management: Licensed shared
  access for 5g,'' \emph{Telecommunications Policy}, vol.~41, no. 5-6, pp.
  422--433, 2017.

\bibitem{santana2018cognitive}
G.~Santana, R.~S. Cristo, C.~Dezan, J.-P. Diguet, D.~P. Osorio, and K.~R.
  Branco, ``Cognitive radio for uav communications: Opportunities and future
  challenges,'' in \emph{2018 International Conference on Unmanned Aircraft
  Systems (ICUAS)}.\hskip 1em plus 0.5em minus 0.4em\relax IEEE, 2018, pp.
  760--768.

\bibitem{zhang2018spectrum}
L.~Zhang, Y.-C. Liang, and M.~Xiao, ``Spectrum sharing for internet of things:
  A survey,'' \emph{IEEE Wireless Communications}, vol.~26, no.~3, pp.
  132--139, 2018.

\bibitem{bayhan2019smart}
S.~Bayhan, A.~Zubow, P.~Gaw{\l}owicz, and A.~Wolisz, ``Smart contracts for
  spectrum sensing as a service,'' \emph{IEEE Transactions on Cognitive
  Communications and Networking}, vol.~5, no.~3, pp. 648--660, 2019.

\bibitem{shen2019uav}
F.~Shen, G.~Ding, Z.~Wang, and Q.~Wu, ``Uav-based 3d spectrum sensing in
  spectrum-heterogeneous networks,'' \emph{IEEE Transactions on Vehicular
  Technology}, vol.~68, no.~6, pp. 5711--5722, 2019.

\bibitem{liu2018spectrum}
X.~Liu, M.~Guan, X.~Zhang, and H.~Ding, ``Spectrum sensing optimization in an
  uav-based cognitive radio,'' \emph{IEEE Access}, vol.~6, pp.
  44\,002--44\,009, 2018.

\bibitem{huang2018cognitive}
Y.~Huang, J.~Xu, L.~Qiu, and R.~Zhang, ``Cognitive uav communication via joint
  trajectory and power control,'' in \emph{2018 IEEE 19th international
  workshop on signal processing advances in wireless communications
  (SPAWC)}.\hskip 1em plus 0.5em minus 0.4em\relax IEEE, 2018, pp. 1--5.

\bibitem{huang2019cognitive}
Y.~Huang, W.~Mei, J.~Xu, L.~Qiu, and R.~Zhang, ``Cognitive uav communication
  via joint maneuver and power control,'' \emph{IEEE Transactions on
  Communications}, vol.~67, no.~11, pp. 7872--7888, 2019.

\bibitem{chen2019interference}
J.~Chen, Y.~Xu, Q.~Wu, Y.~Zhang, X.~Chen, and N.~Qi, ``Interference-aware
  online distributed channel selection for multicluster fanet: A potential game
  approach,'' \emph{IEEE Transactions on Vehicular Technology}, vol.~68, no.~4,
  pp. 3792--3804, 2019.

\bibitem{xu2018efficient}
W.~Xu, S.~Wang, S.~Yan, and J.~He, ``An efficient wideband spectrum sensing
  algorithm for unmanned aerial vehicle communication networks,'' \emph{IEEE
  Internet of Things Journal}, vol.~6, no.~2, pp. 1768--1780, 2018.

\bibitem{qiu2019blockchain}
J.~Qiu, D.~Grace, G.~Ding, J.~Yao, and Q.~Wu, ``Blockchain-based secure
  spectrum trading for unmanned-aerial-vehicle-assisted cellular networks: An
  operator’s perspective,'' \emph{IEEE Internet of Things Journal}, vol.~7,
  no.~1, pp. 451--466, 2019.

\bibitem{hu2018uav}
Z.~Hu, Z.~Zheng, L.~Song, T.~Wang, and X.~Li, ``Uav offloading: Spectrum
  trading contract design for uav-assisted cellular networks,'' \emph{IEEE
  Transactions on Wireless Communications}, vol.~17, no.~9, pp. 6093--6107,
  2018.

\bibitem{azari2020uav}
M.~M. Azari, G.~Geraci, A.~Garcia-Rodriguez, and S.~Pollin, ``Uav-to-uav
  communications in cellular networks,'' \emph{IEEE Transactions on Wireless
  Communications}, vol.~19, no.~9, pp. 6130--6144, 2020.

\bibitem{ghasemi2024accelerating}
A.~Ghasemi and P.~Guinand, ``Accelerating radio spectrum regulation workflows
  with large language models (llms),'' \emph{arXiv preprint arXiv:2403.17819},
  2024.

\bibitem{kuwertz2018applying}
A.~Kuwertz, D.~M{\"u}hlenberg, J.~Sander, and W.~M{\"u}ller, ``Applying
  knowledge-based reasoning for information fusion in intelligence,
  surveillance, and reconnaissance,'' in \emph{Multisensor Fusion and
  Integration in the Wake of Big Data, Deep Learning and Cyber Physical System:
  An Edition of the Selected Papers from the 2017 IEEE International Conference
  on Multisensor Fusion and Integration for Intelligent Systems (MFI
  2017)}.\hskip 1em plus 0.5em minus 0.4em\relax Springer, 2018, pp. 119--139.

\bibitem{maharani2020sentiment}
W.~Maharani, ``Sentiment analysis during jakarta flood for emergency responses
  and situational awareness in disaster management using bert,'' in \emph{2020
  8th International Conference on Information and Communication Technology
  (ICoICT)}.\hskip 1em plus 0.5em minus 0.4em\relax IEEE, 2020, pp. 1--5.

\bibitem{goecks2023disasterresponsegpt}
V.~G. Goecks and N.~R. Waytowich, ``Disasterresponsegpt: Large language models
  for accelerated plan of action development in disaster response scenarios,''
  \emph{arXiv preprint arXiv:2306.17271}, 2023.

\bibitem{lee2023ai}
M.~Lee, L.~Mesicek, K.~Bae, and H.~Ko, ``Ai advisor platform for disaster
  response based on big data,'' \emph{Concurrency and Computation: Practice and
  Experience}, vol.~35, no.~16, p. e6215, 2023.

\bibitem{she2021efficiency}
R.~She and Y.~Ouyang, ``Efficiency of uav-based last-mile delivery under
  congestion in low-altitude air,'' \emph{Transportation Research Part C:
  Emerging Technologies}, vol. 122, p. 102878, 2021.

\bibitem{asadzadeh2022uav}
S.~Asadzadeh, W.~J. de~Oliveira, and C.~R. de~Souza~Filho, ``Uav-based remote
  sensing for the petroleum industry and environmental monitoring:
  State-of-the-art and perspectives,'' \emph{Journal of Petroleum Science and
  Engineering}, vol. 208, p. 109633, 2022.

\bibitem{nova2023ai}
K.~Nova, ``Ai-enabled water management systems: an analysis of system
  components and interdependencies for water conservation,'' \emph{Eigenpub
  Review of Science and Technology}, vol.~7, no.~1, pp. 105--124, 2023.

\bibitem{mashala2023systematic}
M.~J. Mashala, T.~Dube, B.~T. Mudereri, K.~K. Ayisi, and M.~R. Ramudzuli, ``A
  systematic review on advancements in remote sensing for assessing and
  monitoring land use and land cover changes impacts on surface water resources
  in semi-arid tropical environments,'' \emph{Remote Sensing}, vol.~15, no.~16,
  p. 3926, 2023.

\bibitem{adu2017water}
K.~S. Adu-Manu, C.~Tapparello, W.~Heinzelman, F.~A. Katsriku, and J.-D.
  Abdulai, ``Water quality monitoring using wireless sensor networks: Current
  trends and future research directions,'' \emph{ACM Transactions on Sensor
  Networks (TOSN)}, vol.~13, no.~1, pp. 1--41, 2017.

\bibitem{stephenson2019integrating}
P.~J. Stephenson, ``Integrating remote sensing into wildlife monitoring for
  conservation,'' \emph{Environmental Conservation}, vol.~46, no.~3, pp.
  181--183, 2019.

\bibitem{chanev2023application}
M.~Chanev, N.~Dolapchiev, I.~Kamenova, and L.~Filchev, ``Application of remote
  sensing methods for monitoring wild life populations: a review,'' in
  \emph{Ninth International Conference on Remote Sensing and Geoinformation of
  the Environment (RSCy2023)}, vol. 12786.\hskip 1em plus 0.5em minus
  0.4em\relax SPIE, 2023, pp. 731--737.

\bibitem{fourati2021artificial}
F.~Fourati and M.-S. Alouini, ``Artificial intelligence for satellite
  communication: A review,'' \emph{Intelligent and Converged Networks}, vol.~2,
  no.~3, pp. 213--243, 2021.

\bibitem{10473893}
L.~J. Wan, Y.~Huang, Y.~Li, H.~Ye, J.~Wang, X.~Zhang, and D.~Chen, ``Invited
  paper: Software/hardware co-design for llm and its application for design
  verification,'' in \emph{2024 29th Asia and South Pacific Design Automation
  Conference (ASP-DAC)}, 2024, pp. 435--441.

\bibitem{yang2023harnessing}
J.~Yang, H.~Jin, R.~Tang, X.~Han, Q.~Feng, H.~Jiang, S.~Zhong, B.~Yin, and
  X.~Hu, ``Harnessing the power of llms in practice: A survey on chatgpt and
  beyond,'' \emph{ACM Transactions on Knowledge Discovery from Data}, 2023.

\bibitem{javaid2023communication}
S.~Javaid, N.~Saeed, Z.~Qadir, H.~Fahim, B.~He, H.~Song, and M.~Bilal,
  ``Communication and control in collaborative uavs: Recent advances and future
  trends,'' \emph{IEEE Transactions on Intelligent Transportation Systems},
  2023.

\bibitem{ma2023llm}
X.~Ma, G.~Fang, and X.~Wang, ``Llm-pruner: On the structural pruning of large
  language models,'' \emph{Advances in neural information processing systems},
  vol.~36, pp. 21\,702--21\,720, 2023.

\bibitem{xu2024cached}
M.~Xu, D.~Niyato, H.~Zhang, J.~Kang, Z.~Xiong, S.~Mao, and Z.~Han, ``Cached
  model-as-a-resource: Provisioning large language model agents for edge
  intelligence in space-air-ground integrated networks,'' \emph{arXiv preprint
  arXiv:2403.05826}, 2024.

\bibitem{kwon2023efficient}
W.~Kwon, Z.~Li, S.~Zhuang, Y.~Sheng, L.~Zheng, C.~H. Yu, J.~Gonzalez, H.~Zhang,
  and I.~Stoica, ``Efficient memory management for large language model serving
  with pagedattention,'' in \emph{Proceedings of the 29th Symposium on
  Operating Systems Principles}, 2023, pp. 611--626.

\bibitem{chen2024comprehensive}
H.~Chen, J.~Zhang, Y.~Du, S.~Xiang, Z.~Yue, N.~Zhang, Y.~Cai, and Z.~Zhang, ``A
  comprehensive evaluation of fpga-based spatial acceleration of llms,'' in
  \emph{Proceedings of the 2024 ACM/SIGDA International Symposium on Field
  Programmable Gate Arrays}, 2024, pp. 185--185.

\bibitem{huang2024leveraging}
Y.~Huang, ``Leveraging large language models for enhanced nlp task performance
  through knowledge distillation and optimized training strategies,''
  \emph{arXiv preprint arXiv:2402.09282}, 2024.

\bibitem{hassan2023satellite}
S.~S. Hassan, Y.~M. Park, Y.~K. Tun, W.~Saad, Z.~Han, and C.~S. Hong,
  ``Satellite-based its data offloading \& computation in 6g networks: A
  cooperative multi-agent proximal policy optimization drl with attention
  approach,'' \emph{IEEE Transactions on Mobile Computing}, 2023.

\bibitem{chen2024survey}
Q.~Chen, Z.~Guo, W.~Meng, S.~Han, C.~Li, and T.~Q. Quek, ``A survey on resource
  management in joint communication and computing-embedded sagin,'' \emph{arXiv
  preprint arXiv:2403.17400}, 2024.

\bibitem{telli2023comprehensive}
K.~Telli, O.~Kraa, Y.~Himeur, A.~Ouamane, M.~Boumehraz, S.~Atalla, and
  W.~Mansoor, ``A comprehensive review of recent research trends on unmanned
  aerial vehicles (uavs),'' \emph{Systems}, vol.~11, no.~8, p. 400, 2023.

\bibitem{mishra2023autonomous}
S.~Mishra and P.~Palanisamy, ``Autonomous advanced aerial mobility—an
  end-to-end autonomy framework for uavs and beyond,'' \emph{IEEE Access},
  vol.~11, pp. 136\,318--136\,349, 2023.

\bibitem{wang2024generative}
Z.~Wang, J.~Zhang, H.~Du, R.~Zhang, D.~Niyato, B.~Ai, and K.~B. Letaief,
  ``Generative ai agent for next-generation mimo design: Fundamentals,
  challenges, and vision,'' \emph{arXiv preprint arXiv:2404.08878}, 2024.

\bibitem{yao2024survey}
Y.~Yao, J.~Duan, K.~Xu, Y.~Cai, Z.~Sun, and Y.~Zhang, ``A survey on large
  language model (llm) security and privacy: The good, the bad, and the ugly,''
  \emph{High-Confidence Computing}, p. 100211, 2024.

\bibitem{wu2024new}
F.~Wu, N.~Zhang, S.~Jha, P.~McDaniel, and C.~Xiao, ``A new era in llm security:
  Exploring security concerns in real-world llm-based systems,'' \emph{arXiv
  preprint arXiv:2402.18649}, 2024.

\bibitem{wu2023autogen}
Q.~Wu, G.~Bansal, J.~Zhang, Y.~Wu, S.~Zhang, E.~Zhu, B.~Li, L.~Jiang, X.~Zhang,
  and C.~Wang, ``Autogen: Enabling next-gen llm applications via multi-agent
  conversation framework,'' \emph{arXiv preprint arXiv:2308.08155}, 2023.

\bibitem{chacko2023paradigm}
N.~Chacko and V.~Chacko, ``Paradigm shift presented by large language models
  (llm) in deep learning,'' \emph{ADVANCES IN EMERGING COMPUTING TECHNOLOGIES},
  vol.~40, 2023.

\bibitem{elmossallamy2020reconfigurable}
M.~A. ElMossallamy, H.~Zhang, L.~Song, K.~G. Seddik, Z.~Han, and G.~Y. Li,
  ``Reconfigurable intelligent surfaces for wireless communications:
  Principles, challenges, and opportunities,'' \emph{IEEE Transactions on
  Cognitive Communications and Networking}, vol.~6, no.~3, pp. 990--1002, 2020.

\bibitem{faisal2022machine}
K.~Faisal and W.~Choi, ``Machine learning approaches for reconfigurable
  intelligent surfaces: A survey,'' \emph{IEEE Access}, vol.~10, pp.
  27\,343--27\,367, 2022.

\bibitem{yuan2021reconfigurable}
X.~Yuan, Y.-J.~A. Zhang, Y.~Shi, W.~Yan, and H.~Liu,
  ``Reconfigurable-intelligent-surface empowered wireless communications:
  Challenges and opportunities,'' \emph{IEEE wireless communications}, vol.~28,
  no.~2, pp. 136--143, 2021.

\bibitem{de2023llm}
I.~de~Zarz{\`a}, J.~de~Curt{\`o}, G.~Roig, and C.~T. Calafate, ``Llm adaptive
  pid control for b5g truck platooning systems,'' \emph{Sensors}, vol.~23,
  no.~13, p. 5899, 2023.

\bibitem{ali2021urllc}
R.~Ali, Y.~B. Zikria, A.~K. Bashir, S.~Garg, and H.~S. Kim, ``Urllc for 5g and
  beyond: Requirements, enabling incumbent technologies and network
  intelligence,'' \emph{IEEE Access}, vol.~9, pp. 67\,064--67\,095, 2021.

\bibitem{hong2023llm}
Y.~Hong, J.~Wu, and R.~Morello, ``Llm-twin: Mini-giant model-driven beyond 5g
  digital twin networking framework with semantic secure communication and
  computation,'' \emph{arXiv preprint arXiv:2312.10631}, 2023.

\bibitem{lin2023pushing}
Z.~Lin, G.~Qu, Q.~Chen, X.~Chen, Z.~Chen, and K.~Huang, ``Pushing large
  language models to the 6g edge: Vision, challenges, and opportunities,''
  \emph{arXiv preprint arXiv:2309.16739}, 2023.

\bibitem{liang2023unleashing}
Z.~Liang, J.~Cheng, R.~Yang, H.~Ren, Z.~Song, D.~Wu, X.~Qian, T.~Li, and
  Y.~Shi, ``Unleashing the potential of llms for quantum computing: A study in
  quantum architecture design,'' \emph{arXiv preprint arXiv:2307.08191}, 2023.

\bibitem{zhu2015using}
J.~Zhu, A.~Pande, P.~Mohapatra, and J.~J. Han, ``Using deep learning for energy
  expenditure estimation with wearable sensors,'' in \emph{2015 17th
  International Conference on E-health Networking, Application \& Services
  (HealthCom)}.\hskip 1em plus 0.5em minus 0.4em\relax IEEE, 2015, pp.
  501--506.

\bibitem{xia2023towards}
Y.~Xia, M.~Shenoy, N.~Jazdi, and M.~Weyrich, ``Towards autonomous system:
  flexible modular production system enhanced with large language model
  agents,'' in \emph{2023 IEEE 28th International Conference on Emerging
  Technologies and Factory Automation (ETFA)}.\hskip 1em plus 0.5em minus
  0.4em\relax IEEE, 2023, pp. 1--8.

\bibitem{xu2024penetrative}
H.~Xu, L.~Han, Q.~Yang, M.~Li, and M.~Srivastava, ``Penetrative ai: Making llms
  comprehend the physical world,'' in \emph{Proceedings of the 25th
  International Workshop on Mobile Computing Systems and Applications}, 2024,
  pp. 1--7.

\bibitem{formosa2020predicting}
N.~Formosa, M.~Quddus, S.~Ison, M.~Abdel-Aty, and J.~Yuan, ``Predicting
  real-time traffic conflicts using deep learning,'' \emph{Accident Analysis \&
  Prevention}, vol. 136, p. 105429, 2020.

\bibitem{shi2021towards}
L.~Shi, B.~Li, C.~Kim, P.~Kellnhofer, and W.~Matusik, ``Towards real-time
  photorealistic 3d holography with deep neural networks,'' \emph{Nature}, vol.
  591, no. 7849, pp. 234--239, 2021.

\bibitem{srar2010adaptive}
J.~A. Srar, K.-S. Chung, and A.~Mansour, ``Adaptive array beamforming using a
  combined lms-lms algorithm,'' \emph{IEEE Transactions on Antennas and
  Propagation}, vol.~58, no.~11, pp. 3545--3557, 2010.

\bibitem{bariah2023large}
L.~Bariah, Q.~Zhao, H.~Zou, Y.~Tian, F.~Bader, and M.~Debbah, ``Large language
  models for telecom: The next big thing?'' \emph{arXiv preprint
  arXiv:2306.10249}, 2023.

\bibitem{pathirana2018towards}
D.~Pathirana, \emph{Towards better regulation of unmanned aerial vehicles in
  national airspace: a comparative analysis of selected national
  regulations}.\hskip 1em plus 0.5em minus 0.4em\relax McGill University
  (Canada), 2018.

\bibitem{9537930}
W.~Y.~B. Lim, S.~Garg, Z.~Xiong, Y.~Zhang, D.~Niyato, C.~Leung, and C.~Miao,
  ``Uav-assisted communication efficient federated learning in the era of the
  artificial intelligence of things,'' \emph{IEEE Network}, vol.~35, no.~5, pp.
  188--195, 2021.

\end{thebibliography}

\end{document}